\PassOptionsToPackage{table,xcdraw}{xcolor}

\documentclass{bmvc2k}

\hypersetup{colorlinks=false, hidelinks}

\usepackage{url}
\usepackage{microtype} \usepackage[utf8]{inputenc}

\usepackage{graphicx}
\usepackage{amsmath,amssymb}
\usepackage{booktabs}
\usepackage{multirow}
\usepackage{xspace}
\usepackage{subcaption}
\usepackage{float} \usepackage{placeins} \usepackage{needspace} 

\usepackage{enumitem}
\usepackage[normalem]{ulem}
    \useunder{\uline}{\ul}{}

\usepackage{caption}
\usepackage{rotating}
\usepackage{pdflscape}

\newcommand{\method}{\textit{OrthoPairGTL}\xspace}
\newcommand{\opa}{\textit{OPA}\xspace}
\newcommand{\gtl}{\textit{GTL}\xspace}

\newcommand{\RtoF}{Real$\!\to\!$Fake\xspace}
\newcommand{\RtoR}{Real$\!\to\!$Real\xspace}
\newcommand{\rfpair}{real--fake\xspace}

\makeatletter
\newcommand{\defFloatSpacing}[1]{\@ifundefined{floatspace@#1@before}{\expandafter\newlength\csname floatspace@#1@before\endcsname \expandafter\newlength\csname floatspace@#1@capAbove\endcsname \expandafter\newlength\csname floatspace@#1@capBelow\endcsname \expandafter\newlength\csname floatspace@#1@after\endcsname }{}}
\newcommand{\setFloatSpacing}[3]{\expandafter\setlength\csname floatspace@#1@#2\endcsname{#3}}
\newcommand{\fsBefore}[1]{\vspace{\csname floatspace@#1@before\endcsname}}
\newcommand{\fsCapAbove}[1]{\vspace{\csname floatspace@#1@capAbove\endcsname}}
\newcommand{\fsCapBelow}[1]{\vspace{\csname floatspace@#1@capBelow\endcsname}}
\newcommand{\fsAfter}[1]{\vspace{\csname floatspace@#1@after\endcsname}}
\makeatother

\defFloatSpacing{figPipeline}
\defFloatSpacing{tabIndomain}
\defFloatSpacing{tabCrossResults}
\defFloatSpacing{figLoraDynamics}
\defFloatSpacing{tabAblation}

\setFloatSpacing{figPipeline}{capAbove}{8pt}

\makeatletter
\renewcommand\section{\@startsection {section}{1}{\z@}{-2.3ex \@plus -0.8ex \@minus -.2ex}{1.4ex \@plus.2ex}{\normalfont\Large\bfseries\flushleft\textcolor{bmv@sectioncolor}}}
\renewcommand\subsection{\@startsection{subsection}{2}{\z@}{-2.0ex\@plus -0.8ex \@minus -.2ex}{0.8ex \@plus .2ex}{\normalfont\large\bfseries\flushleft\textcolor{bmv@sectioncolor}}}
\makeatother
\AtBeginDocument{\setlength{\abovedisplayskip}{6pt plus 2pt minus 3pt}\setlength{\belowdisplayskip}{6pt plus 2pt minus 3pt}\setlength{\abovedisplayshortskip}{3pt plus 1pt minus 2pt}\setlength{\belowdisplayshortskip}{3pt plus 1pt minus 2pt}}

\title{Paired Geometric Supervision for Generalizable Deepfake Detection}

\addauthor{Luke Collins}{l.collins@deakin.edu.au}{1}
\addauthor{Chang-Tsun Li}{chang-tsun.li@deakin.edu.au}{1}
\addauthor{Duc Thanh Nguyen}{duc.nguyen@deakin.edu.au}{1}
\addauthor{Victor Sanchez}{v.f.sanchez-silva@warwick.ac.uk}{2,3}

\addinstitution{
 School of Information Technology\\
 Deakin University\\
 Waurn Ponds, Australia
}
\addinstitution{
 Department of Computer Science\\
 University of Warwick\\
 Coventry, UK
}

\makeatletter
\def\bmv@authornote{$^{3}$Victor Sanchez is also with Universitat Aut\`onoma de Barcelona, Spain, as a Distinguished Researcher. This work was partially funded by the Spanish Ministry of Science, Innovation and Universities (MICIU) -- ATR2024-154513.}
\makeatother

\runninghead{Collins \etal}{Paired Geometric Supervision}

\def\etal{\emph{et al}\bmvaOneDot}

\renewcommand{\arraystretch}{0.96}

\newcommand{\FigTwoWidth}{0.80\linewidth}

\begin{document}

\maketitle

\begin{abstract}
We present \method, a geometric training framework for cross-domain deepfake detection, built on a partially frozen DINOv3 backbone ($\sim$300M parameters), and make three contributions:
\textbf{(1)} We re-evaluate six recent transfer-learning detectors (Forensics Adapter, Effort, three GenD variants and FakeSTormer) using the authors' released checkpoints under a single preprocessing pipeline across 17 cross-domain datasets (2018--2025), so that the reported gaps reflect method rather than per-method pipeline tuning.
\textbf{(2)} We model the transformation between paired real and fake samples, training Parameter-Efficient Fine-Tuning (PEFT) adapters with an orthogonal pair decomposition, hyperspherical regularisation, and (in the GSD-LoRA variant) a set-level coherence objective.
\textbf{(3)} Under LoRA, this regularisation gives a more stable optimisation profile: GSD-LoRA's post-peak validation degradation is $4.2{\times}$ shallower than plain LoRA at the same parameter footprint, and the time-averaged validation AUC across training is 1.82\,pp higher.
On the 17 datasets, GSD-LoRA matches or exceeds every re-evaluated baseline (92.82\% vs.\ 91.58\% for the strongest; paired Wilcoxon signed-rank $p{=}0.015$, 13/17 per-dataset wins), with the largest gains on modern reenactment-heavy benchmarks (DSv1.1, DSv2). Under strict parameter budgets, LayerNorm-only tuning ($\approx$0.03\% of the backbone) is the most stable configuration and matches the strongest baseline with a mean of 91.97\%.
\end{abstract}
 
\fsBefore{figPipeline}
\begin{figure}[h!]
\centering
\includegraphics[width=0.9\linewidth]{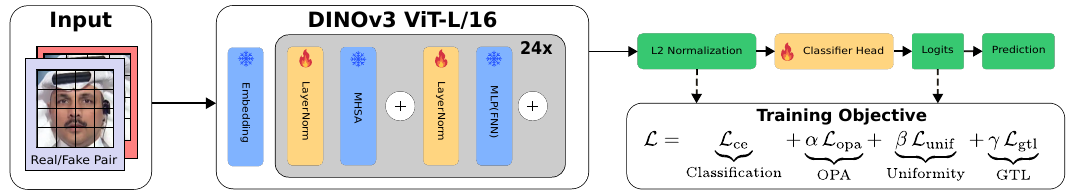}
\fsCapAbove{figPipeline}\caption{\textbf{Overview of \method.} A frozen DINOv3 backbone encodes temporally aligned \rfpair pairs. Orthogonal Paired Alignment ($\mathcal{L}_{\text{opa}}$) decomposes these embeddings to isolate manipulation artefacts at the frame level. \textbf{Dashed lines} denote auxiliary geometric objectives, including Geodesic Transformation Learning ($\mathcal{L}_{\text{gtl}}$) and Set-Geometric Descriptor (GSD) regularisation, which structure both the frame-level and video-level training spaces, but are discarded at inference.}
\label{fig:pipeline}
\fsCapBelow{figPipeline}
\end{figure}
\fsAfter{figPipeline}

\section{Introduction}
\label{sec:introduction}
Generative models now produce synthetic images and videos that are difficult to distinguish from real ones. Detectors trained on a single domain perform well on familiar test sets such as FaceForensics++~\cite{roessler2019faceforensics++} but degrade on unseen manipulation types, compression settings, or source distributions~\cite{yan2023deepfakebench}. This \emph{generalisation gap} is the primary obstacle to practical deployment. Pretrained encoders offer a stronger starting point than training from scratch: backbones such as DINOv3 and CLIP variants provide transferable features that downstream detectors can adapt with limited supervision. Most current methods, however, treat real and fake samples as independent classes, discarding the paired structure and encouraging models to rely on dataset-specific correlations such as background statistics or sensor noise rather than on the manipulation artefacts themselves. Robust cross-domain detection requires modelling the transformation between real and fake, not just the class label. Paired training data, temporally and contextually aligned real and fake frames, lets us control for confounds such as pose, lighting, and background; what remains between the two frames is the manipulation, which becomes the discriminative signal~\cite{yermakov2026deepfake}. Existing paired methods typically compare raw features and miss the shared geometric structure of manipulations.
\method addresses this by structuring the embedding space of frozen foundation models through constraints on paired \RtoF displacement vectors (Fig.~\ref{fig:pipeline}). Rather than allowing the adapter to learn arbitrary feature shifts, we enforce geometric constraints that isolate manipulation artefacts along consistent, identity-orthogonal trajectories at both the frame and video levels. We introduce four such constraints:

\begin{enumerate}[topsep=2pt, itemsep=2pt, parsep=0pt, partopsep=0pt]
\item \textbf{Orthogonal Paired Alignment (\opa):} We decompose \rfpair pairs into a manipulation direction and an identity-preserving residual. Enforcing orthogonality concentrates forensic evidence along specific axes while suppressing identity leakage.
\item \textbf{Geodesic Transformation Learning (\gtl):} Instead of binary classification alone, we train a classifier to distinguish malicious \RtoF trajectories from benign \RtoR variations using a geodesic descriptor.
\item \textbf{Manifold Regularisation:} To prevent feature collapse and improve coverage on the hypersphere, we apply a hyperspherical uniformity objective to the batch embeddings and to the orthogonal residuals of each pair.
\item \textbf{Set-Geometric Descriptor (GSD) Regularisation:} We enforce video-level consistency through two within-video terms, a directional concentration of fake displacements ($\mathcal{L}_{\text{conc}}$) and residual coherence across frames ($\mathcal{L}_{\text{coh}}$), plus a smaller between-video dispersion penalty ($\mathcal{L}_{\text{disp}}$) whose effect is expected to be modest under our default video-grouped sampler (see Sec.~\ref{subsec:gsd}).
\end{enumerate}

\noindent We freeze the DINOv3 backbone and adapt it with lightweight Parameter-Efficient Fine-Tuning (PEFT) modules and a small classification head to limit over-fitting. Our contributions are:

\begin{itemize}[topsep=2pt, itemsep=2pt, parsep=0pt, partopsep=0pt]
\item \textbf{Single-pipeline cross-domain evaluation.} We re-evaluate six recent transfer-learning baselines using the authors' released checkpoints under a single unified preprocessing pipeline across 17 datasets (2018--2025), separating method gains from per-method pipeline tuning and providing a reproducible reference for future work.
\item \textbf{Training stability under PEFT.} Combining frame-level pair geometry with set-level coherence regularisation produces a $4.2{\times}$ shallower post-peak validation degradation slope than plain LoRA at the same parameter footprint (Table~\ref{tab:training_stability}), broadening the high-validation plateau.
\item \textbf{Reenactment-specific cross-domain wins.} Our largest per-dataset advantages appear on the speech-driven reenactment benchmarks (DSv1.1, DSv2), where per-frame manipulation cues are weakest; on RGB face-swap benchmarks LayerNorm-only tuning remains competitive.
\item \textbf{Method.} We formulate detection as learning the \RtoF transformation between paired samples using an orthogonal pair decomposition (distinct from prior subspace approaches such as Effort~\cite{yan2025orthogonal}), hyperspherical uniformity, and a Set-Geometric Descriptor (GSD) in the GSD-LoRA variant.
\end{itemize}
 \section{Related Work}
\label{sec:related}
\paragraph{Generalisable Deepfake Detection.}
Reliability under dataset shift is a central problem in multimedia forensics: strong in-benchmark performance often fails to transfer. Recent work uses foundation models and tracks progress on benchmarking frameworks such as DeepfakeBench~\cite{yan2023deepfakebench}. Two common strategies have emerged: architectural adaptation and representation-level augmentation. Forensics Adapter~\cite{cui2025forensics} illustrates architectural adaptation by adding a specialised side network on top of CLIP to capture boundary artefacts, thereby improving transferability without full model retraining. On the augmentation side, LSDA~\cite{yan2024transcending} diversifies the training distribution by simulating latent variations and interpolations, reducing forgery-specific overfitting. Our work is complementary: rather than synthesising feature-space samples, we constrain the geometry of aligned \rfpair embeddings, so the representation encodes coherent \RtoF transformations rather than invariance to unconstrained perturbations. Our approach is closest to GenD~\cite{yermakov2026deepfake}, which uses paired supervision to reduce shortcut learning and improve cross-dataset generalisation.
\vspace*{-2ex}
\paragraph{Parameter-Efficient Fine-Tuning (PEFT).}
Adapting large pretrained encoders with minimal updates is computationally efficient but carries risks in forensic settings. LoRA~\cite{hu2022lora} and other expressive adapters can improve in-domain accuracy but may degrade cross-domain robustness if the learned representation concentrates on a narrow set of repetitive cues. Recent work on orthogonal subspace decomposition points to an \emph{asymmetry phenomenon} in AI-generated image detection: models fit quickly to seen fake patterns while under-representing the diversity of real data, thereby hurting generalisation~\cite{yan2025orthogonal}. To address this, Effort~\cite{yan2025orthogonal} preserves pretrained semantic structure while learning forgery cues in an orthogonal residual subspace. We adopt PEFT settings (LayerNorm tuning as the primary configuration, LoRA as a secondary option) and add paired geometric constraints to regularise adaptation, encouraging identity-consistent structure while modelling manipulation-specific displacement. Concurrent with our submission, SICA~\cite{du2026can} also constrains low-rank adaptation of a frozen ViT so that the pretrained semantic manifold is preserved while artefact features are learned, targeting a single monolithic detector across forensic subdomains; our constraints are instead pairwise and geometric, shaping the \RtoF displacement rather than a subdomain-unified artefact space. Its image-level, aggregate-trained protocol is not comparable to our train-on-FF++-only video evaluation, so we discuss rather than tabulate it.
\paragraph{Geometry and Subspace Decomposition.}
Imposing geometric structure on learned representations can improve transfer. Hyperspherical uniformity, for example, is associated with more transferable representations~\cite{wang2020understanding}. The orthogonal subspace perspective~\cite{yan2025orthogonal} offers a complementary view by decomposing model updates to preserve pretrained semantics while learning forgery-specific components. We shift from a \emph{global} decomposition of model weights to a \emph{local} decomposition in embedding space. For each aligned \rfpair, we estimate an instance-specific manipulation direction and an orthogonal context-preserving residual that captures factors common to the pair (e.g., pose, illumination, and background). The semantic role of these components depends on the type of manipulation: for identity-preserving edits (e.g., reenactment), the residual typically retains the subject identity under a matched context. For identity-changing edits (e.g., face swapping), the identity shift is absorbed into the \RtoF transformation vector, and the orthogonal residual retains the shared scene context and comparable manipulation cues. This often produces larger and more coherent displacement vectors for face swaps, making manipulation \emph{types} more separable, whereas reenactment induces subtler displacements that concentrate around identity-related structure. The formulation casts detection as a transformation learning problem (\RtoF), motivating our \opa objective and \gtl-based transformation discrimination. \section{Methodology}
\label{sec:methodology}
\setlength{\abovedisplayskip}{4pt}
\setlength{\belowdisplayskip}{4pt}
\setlength{\abovedisplayshortskip}{2pt}
\setlength{\belowdisplayshortskip}{2pt}

We propose \method, which reformulates deepfake detection as a geometric transformation problem: instead of learning a static decision boundary between real and fake samples, we model the \emph{displacement} induced by manipulation between paired samples on a hyperspherical representation manifold.

\subsection{Preliminaries and Parameter-Efficient Tuning}
\label{subsec:setup}
Let $\mathcal{D} = \{(\mathbf{x}_r, \mathbf{x}_f)\}$ be a dataset of paired face frames, where $\mathbf{x}_r$ is a pristine frame and $\mathbf{x}_f$ is its manipulated counterpart sharing the same identity and scene context. A backbone encoder $f_\theta$ maps an input $\mathbf{x}$ to a feature vector $\mathbf{u} := f_\theta(\mathbf{x}) \in \mathbb{R}^d$. We use the DINOv3 Vision Transformer~\cite{simeoni2025dinov3} as the foundation model and project its features onto the unit hypersphere for geometric stability:
\begin{equation}
\label{eq:hypersphere}
\mathbf{z} = \frac{\mathbf{u}}{\|\mathbf{u}\|_2} \in \mathbb{S}^{d-1}.
\end{equation}
A lightweight classifier $g(\cdot)$ predicts real vs.\ fake from $\mathbf{u}$ (or $\mathbf{z}$) and is optimised with cross-entropy loss $\mathcal{L}_{\text{ce}}$. Mini-batches preserve \rfpair pairing when available. To adapt a large-scale backbone (${\sim}300$M parameters) for forensics under a strict training budget, we evaluate two parameter-efficient strategies.

\paragraph{Primary Configuration: LayerNorm Tuning (LN).}
Following recent parameter-efficient baselines~\cite{yermakov2026deepfake,demin2023effectiveness,giannou2023expressive}, our primary setting freezes the backbone and updates only the affine parameters (scale and shift) of LayerNorm layers, along with a lightweight classification head. On the order of $10^5$ parameters are trained (${\approx}0.03\%$ of the ViT-L/16 backbone), which restricts the capacity available for dataset-specific memorisation and forces reuse of pretrained semantics.

\paragraph{Secondary Configuration: Low-Rank Adaptation (LoRA).}
When a higher adaptation capacity is desired, we attach LoRA modules~\cite{hu2022lora} to every linear projection inside the ViT-L/16 transformer blocks: the four attention projections ($\mathbf{W}_Q, \mathbf{W}_K, \mathbf{W}_V, \mathbf{W}_O$) and both feed-forward projections ($\mathbf{W}_{\text{up}}, \mathbf{W}_{\text{down}}$).
Each adapter learns a rank-$r$ update $\Delta\mathbf{W} = \mathbf{B}\mathbf{A}$ with $r=64$ and scaling factor $\alpha/r = 128/64 = 2$, applied with dropout $0.05$ on the input branch. The backbone is otherwise frozen. Predictions are produced by a two-layer MLP head (hidden dimension $2048$, dropout $0.45$) trained from scratch; the total trainable footprint is $28.3$M backbone parameters ($9.34\%$ of the $303.1$M frozen ViT-L/16) plus a $2.1$M head. Higher-capacity adapters are prone to the \emph{asymmetry phenomenon}~\cite{yan2025orthogonal}, where rapid memorisation degrades generalisation; our geometric constraints (Secs.~\ref{subsec:opa}--\ref{subsec:gsd}) push the additional capacity toward identity-orthogonal manipulation cues rather than dataset-specific correlations. The peak checkpoint is selected via early stopping on a held-out validation suite.

\subsection{Orthogonal Paired Alignment (\opa)}
\label{subsec:opa}
Standard classifiers can exploit identity or background cues that correlate with forgery, which reduces cross-domain robustness. We address this with paired supervision and an explicit pairwise geometry in embedding space. Unlike approaches that decompose a \emph{global} dataset subspace (e.g., Effort~\cite{yan2025orthogonal}), we perform a \emph{local}, instance-specific decomposition for each \rfpair pair. Given a paired and temporally/contextually aligned frame pair $(\mathbf{x}_r,\mathbf{x}_f)$, we obtain their normalised embeddings
$\mathbf{z}_r = \frac{f_\theta(\mathbf{x}_r)}{\|f_\theta(\mathbf{x}_r)\|_2}$ and
$\mathbf{z}_f = \frac{f_\theta(\mathbf{x}_f)}{\|f_\theta(\mathbf{x}_f)\|_2}$ (cf.\ Sec.~\ref{subsec:setup}). We then define the pairwise manipulation direction as:
\begin{equation}
\label{eq:manipulation_dir}
\mathbf{v} = \frac{\mathbf{z}_f - \mathbf{z}_r}{\|\mathbf{z}_f - \mathbf{z}_r\|_2}.
\end{equation}
Each embedding decomposes into a scalar projection along $\mathbf{v}$ and an orthogonal residual using $\epsilon$-stabilised normalisation ($\epsilon{>}0$, a small additive constant):
\begin{equation}
\label{eq:opa_decomp}
\alpha_k = \mathbf{z}_k^\top \mathbf{v}, \qquad
\mathbf{z}_{\perp,k} = \frac{\mathbf{z}_k - \alpha_k \mathbf{v}}{\|\mathbf{z}_k - \alpha_k \mathbf{v}\|_2 + \epsilon},
\quad k\in\{r,f\}.
\end{equation}
\noindent Here $\alpha_k$ is the coordinate of $\mathbf{z}_k$ along the \RtoF direction, and $\mathbf{z}_{\perp,k}$ retains the variation orthogonal to that axis. With a per-pair $\mathbf{v}$, removing the projection onto the displacement direction lands both pair members on the same hyperplane orthogonal to $\mathbf{v}$, so the two residuals coincide identically ($\mathbf{z}_{\perp,r} \equiv \mathbf{z}_{\perp,f}$): the decomposition partitions every pair into a one-dimensional manipulation coordinate and a shared $(d{-}2)$-dimensional residual. The \opa objective enforces a minimum displacement of the fake along $\mathbf{v}$ by margin $m$:
\begin{equation}
\label{eq:opa_loss}
\mathcal{L}_{\text{opa}} = \mathrm{ReLU}(m - (\alpha_f - \alpha_r))^2.
\end{equation}
Since $\mathbf{v}$ is the unit vector along $\mathbf{z}_f - \mathbf{z}_r$, $\alpha_f - \alpha_r$ equals the chord length $\|\mathbf{z}_f - \mathbf{z}_r\|_2$ identically, so $\mathcal{L}_{\text{opa}}$ is a chord-length hinge that prevents paired embeddings from collapsing. Although the within-pair residuals coincide by construction, $\mathbf{z}_\perp$ remains the analysis target downstream: the cross-pair residual uniformity term (Sec.~\ref{subsec:manifold}) organises the orthogonal complement across pairs, shaping the identity-preserving structure verified in Sec.~\ref{subsec:repr} and cross-checked against an independent ArcFace embedding. A per-video-mean alternative that replaces $\mathbf{v}$ with $\bar{\mathbf{v}}$ (its per-video mean) restores a within-pair residual gradient but degrades cross-domain transfer by $-0.68$ MEAN-17 (supp.\ Sec.~\ref{sec:supp_ablation}), confirming the shared-hyperplane construction. In all experiments \opa is computed on paired samples when available, and $m$ is a fixed margin hyperparameter (Sec.~\ref{subsec:protocol}).

\subsection{Manifold Regularisation}
\label{subsec:manifold}
To prevent feature collapse and encourage broad coverage of $\mathbb{S}^{d-1}$, we apply a hyperspherical uniformity objective to the batch embeddings, extended by residual uniformity, which spreads the \opa orthogonal residuals across the batch.
\paragraph{Hyperspherical Uniformity.}
We minimise the log average Gaussian potential~\cite{wang2020understanding}:
\begin{equation}
\label{eq:unif}
\mathcal{L}_{\text{unif}} =
\log \left( \frac{1}{N(N-1)} \sum_{i \neq j} e^{-t \|\mathbf{z}_i - \mathbf{z}_j\|_2^2} \right).
\end{equation}
Here $N$ is the number of batch embeddings and $t{=}2$ is a fixed temperature (the residual variant below uses $t_\perp{=}3.0$).

\paragraph{Residual Uniformity.}
We also apply $\mathcal{L}_{\text{unif}}$ to the concatenated batch of \opa orthogonal residuals $\{\mathbf{z}_{\perp,r,i}, \mathbf{z}_{\perp,f,i}\}_{i=1}^{P}$ from the paired samples in each batch (with a separate temperature $t_\perp = 3.0$). Because $\mathbf{z}_{\perp,r,i} = \mathbf{z}_{\perp,f,i}$ within a pair (Eq.~\ref{eq:opa_decomp}), the within-pair distances contribute trivially and the cross-pair distances spread the residual subspace across the batch, shaping the identity-preserving structure analysed in Sec.~\ref{subsec:repr}.
\subsection{Geodesic Transformation Learning (\gtl)}
\label{subsec:gtl}
\opa structures individual pairs but does not separate malicious manipulation from benign intra-class variation. Geodesic Transformation Learning (\gtl) addresses this by discriminating \RtoF (RF) transformations from \RtoR (RR) variations. For any two embeddings $(\mathbf{z}_a, \mathbf{z}_b)$ --- with $(\mathbf{z}_a,\mathbf{z}_b){=}(\mathbf{z}_r,\mathbf{z}_f)$ for \RtoF and $(\mathbf{z}_r,\mathbf{z}_{r'})$ for \RtoR (Eq.~\ref{eq:hard_rr}) --- we define a compact transformation descriptor:
\begin{equation}
\label{eq:gtl_descriptor}
\begin{aligned}
\mathbf{v}_{ab} &= \tfrac{\mathbf{z}_b-\mathbf{z}_a}{\|\mathbf{z}_b-\mathbf{z}_a\|_2 + \epsilon}, &\quad
\delta_{ab} &= \|\mathbf{z}_b-\mathbf{z}_a\|_2, \\
\psi_{ab} &= \arccos\!\Big(\mathrm{clip}(\mathbf{z}_a^\top\mathbf{z}_b, -1, 1)\Big), &\quad
\mathcal{T}_{ab} &= [\,\mathbf{v}_{ab};\ \delta_{ab};\ \psi_{ab}\,].
\end{aligned}
\end{equation}
The descriptor captures three aspects of the transformation: $\mathbf{v}_{ab}$ encodes the unit direction of displacement, $\delta_{ab}$ the Euclidean magnitude, and $\psi_{ab}$ the angular distance on the hypersphere\footnote{$\psi_{ab}$ is the great-circle geodesic distance on $\mathbb{S}^{d-1}$; the full descriptor also includes the Euclidean chord magnitude and direction, so \gtl is geodesic-aware rather than a fully Riemannian formulation. The angular term is the load-bearing component on identity-swap benchmarks; a strictly Log-map-based variant is left to future work.} A lightweight encoder $E(\cdot)$ maps $\mathcal{T}$ to a low-dimensional representation, and a binary classifier $C(\cdot)$ predicts RF vs.\ RR.
\paragraph{Hard Negative RR Pairing.}
To construct difficult RR pairs, for each real embedding $\mathbf{z}_r$ in the mini-batch we select the most similar \emph{other} real embedding by cosine similarity:
\begin{equation}
\label{eq:hard_rr}
r'=\arg\max_{j\neq r,\ y_j=0}\big(\mathbf{z}_r^\top \mathbf{z}_j\big).
\end{equation}
\noindent The \gtl objective minimises classification error while constraining benign transformations to remain small:
\begin{equation}
\label{eq:gtl_loss}
\mathcal{L}_{\text{gtl}} = \mathcal{L}_{\text{ce}}^{\text{gtl}}\!\big(C(E(\mathcal{T}_{rf})), 1\big) + \mathcal{L}_{\text{ce}}^{\text{gtl}}\!\big(C(E(\mathcal{T}_{rr'})), 0\big) + \lambda_{\text{small}}\big(\psi_{rr'}^2 + c\,\delta_{rr'}^2\big),
\end{equation}
where $\mathcal{L}_{\text{ce}}^{\text{gtl}}$ is the cross-entropy loss for the \gtl classifier (distinct from the main classification loss $\mathcal{L}_{\text{ce}}$).
The first two terms train the classifier to distinguish RF (label 1) from RR (label 0) transformations; the regularisation term (weighted by $\lambda_{\text{small}}$) penalises large RR transformations and keeps benign real--real variation geometrically compact.
The constant $c$ balances the angular and Euclidean components of the penalty ($c=0.01$ throughout).

\subsection{Set-Geometric Descriptors (GSD)}
\label{subsec:gsd}
\opa, \gtl, and the manifold regulariser all act at the \emph{pair} level (a single real--fake tuple). Within a single-source video, however, a manipulation method produces a \emph{coherent} family of transformations: identity-swap methods push features in a similar direction, and reenactment methods produce more varied displacements while still sharing a common residual subspace. We capture this set-level structure with three lightweight regularisers, the Set-Geometric Descriptors, abbreviated GSD rather than SGD to avoid the clash with stochastic gradient descent, computed within a video group $\mathcal{G}=\{(\mathbf{z}_{r,i},\mathbf{z}_{f,i})\}_{i=1}^{K}$ of $K$ paired frames sharing a video identifier. For each pair $i$ let $\mathbf{v}_i=(\mathbf{z}_{f,i}-\mathbf{z}_{r,i})/\|\mathbf{z}_{f,i}-\mathbf{z}_{r,i}\|_2$ be the unit displacement direction and $\hat{\mathbf{z}}_{\perp,i}$ the unit-normalised \opa residual. Two within-video terms drive the per-group loss:
\begin{equation}
\label{eq:gsd_conc_coh}
\mathcal{L}_{\text{conc}}(\mathcal{G}) = -\log\max\!\left(\Big\|\frac{1}{K}\sum_{i=1}^{K}\mathbf{v}_i\Big\|_2,\ \varepsilon\right),
\qquad
\mathcal{L}_{\text{coh}}(\mathcal{G}) = -\frac{2}{K(K-1)}\sum_{i<j}\hat{\mathbf{z}}_{\perp,i}^{\top}\hat{\mathbf{z}}_{\perp,j}.
\end{equation}
The \emph{concentration} term penalises within-video disagreement on the manipulation direction: it is the negative log norm of the mean direction, minimised when the within-video $\mathbf{v}_i$ are aligned; the floor $\varepsilon$ (distinct from the stabiliser $\epsilon$ of Eqs.~\ref{eq:opa_decomp},~\ref{eq:gtl_descriptor}) keeps the log finite. The \emph{coherence} term encourages the residual subspace to be stable across frames of the same video.
A third between-video \emph{dispersion} term operates on per-video mean directions.
Letting $\boldsymbol{\mu}_g = \tfrac{1}{K}\sum_{i}\mathbf{v}_{g,i}$ denote the mean direction of video $g$, we apply the Wang--Isola uniformity criterion~\cite{wang2020understanding} to the set $\{\boldsymbol{\mu}_g\}_{g=1}^{G}$ of per-video mean directions ($G$ videos per batch; distinct from $N$ in Eq.~\ref{eq:unif}):
\begin{equation}
\label{eq:gsd_disp}
\mathcal{L}_{\text{disp}} = \log\!\left(\frac{1}{G^2}\sum_{g,h} e^{-t\|\boldsymbol{\mu}_g-\boldsymbol{\mu}_h\|_2^2}\right),
\end{equation}
which prevents all manipulation fingerprints from collapsing onto a single direction.
Computing GSD requires that the dataloader emit complete real--fake pairs grouped by video; we use a \emph{video-grouped batch sampler} with $K{=}4$ frames per video (Sec.~\ref{subsec:protocol}).
$\mathcal{L}_{\text{disp}}$ is included with a small weight $\eta_{\text{d}} {=} 2{\times}10^{-3}$ in our default configuration, substantially smaller than the within-video terms ($\eta_{\text{c}} {=} \eta_{\text{h}} {=} 5{\times}10^{-3}$). Because the default video-grouped sampler does not stratify batches by manipulation method, $\mathcal{L}_{\text{disp}}$ acts as a weak anti-collapse regulariser; a method-stratified ablation is left to future work.
\paragraph{Where GSD bites.}
GSD becomes the active regulariser only when the adapter has enough capacity to memorise per-frame cues: LN's ${\sim}10^5$ trainable parameters cannot fit such cues in the first place, so set-level coherence is largely redundant, but LoRA increases the trainable budget ${\approx}283\times$. The within-video prior begins to shape which degrees of freedom the adapter uses. The coherence term operates on $\mathbf{z}_\perp$ rather than the full embedding because, by construction (Sec.~\ref{subsec:opa}), $\mathbf{z}_\perp$ isolates the pair-shared scene context (pose, lighting, background, and for reenactment identity). The single-framework ablation in Sec.~\ref{subsec:ablate} (Table~\ref{tab:ablation}) supports this read: under LN's tight budget the geometric losses (uniformity, \gtl, GSD) are individually neutral or harmful. In contrast, under LoRA every geometric addition is beneficial. We refer to the GSD+LoRA combination as GSD-LoRA; its training dynamics are characterised in Fig.~\ref{fig:lora_dynamics}.

\subsection{Overall Objective}
\label{subsec:objective}
We optimise the combined objective with fixed scalar weights that balance the pair-level constraints (\opa, uniformity, \gtl) against the set-level GSD regularisers and the primary classification task:
\begin{equation}
\label{eq:total}
\mathcal{L}_{\text{total}} = \mathcal{L}_{\text{ce}} + \alpha\,\mathcal{L}_{\text{opa}} + \beta\,\mathcal{L}_{\text{unif}} + \gamma\,\mathcal{L}_{\text{gtl}}
+ \eta_{\text{c}}\mathcal{L}_{\text{conc}} + \eta_{\text{h}}\mathcal{L}_{\text{coh}} + \eta_{\text{d}}\mathcal{L}_{\text{disp}}.
\end{equation}
\paragraph{Weight Selection.}
We initially explored uncertainty-based task weighting~\cite{kendall2018multi} to automatically balance the loss components. Still, we found that statically fixed weights gave more stable optimisation and marginally better generalisation: $\alpha{=}1.0$, $\beta{=}0.5$, $\gamma{=}1.0$, the \opa margin $m{=}1.5$, the \gtl small-transform penalty $\lambda_{\text{small}}{=}0.2$ and mixing constant $c{=}0.01$, and the GSD within-video weights $\eta_{\text{c}}{=}\eta_{\text{h}}{=}5\!\times\!10^{-3}$ and smaller between-video dispersion weight $\eta_{\text{d}}{=}2\!\times\!10^{-3}$ (supp.\ Table~\ref{tab:supp_loss_weights} consolidates all values; see Sec.~\ref{subsec:gsd} for the batch-composition caveat that limits $\mathcal{L}_{\text{disp}}$'s expected effect). Table~\ref{tab:ablation} is an on/off component ablation rather than a weight-magnitude study; it shows \gtl ($\gamma$) contributes the most to cross-domain transfer on identity-diverse benchmarks. Complementary single-seed sweeps over the \opa margin $m$, the small-transform penalty $\lambda_{\text{small}}$ and the \gtl mixing constant $c$, each spanning an order of magnitude, all leave cross-domain MEAN-17 within a $0.5$-point band (Supp.\ Sec.~\ref{sec:supp_hyperparam_sweeps}, Table~\ref{tab:supp_hyperparam_sweeps}), so none of these hyperparameters is load-bearing. We hold $\alpha,\beta,\gamma$ and $\eta_{\text{d}}$ fixed rather than magnitude-swept; multi-seed sweeps over all weights are left to future work (compute budget). At inference the geometric branches (encoder $E$ and classifier $C$) are discarded; only the backbone and classification head are used, so the inference cost is that of standard feature extraction and classification.
\paragraph{Train-/test-time objective alignment.}
The auxiliary geometric objectives ($\mathcal{L}_{\text{opa}}$, $\mathcal{L}_{\text{unif}}$, $\mathcal{L}_{\text{gtl}}$, $\mathcal{L}_{\text{conc}}$, $\mathcal{L}_{\text{coh}}$) are regularisers: they shape the backbone's feature manifold during gradient updates and are discarded at inference, in the same spirit as auxiliary-loss multi-task training or contrastive pretraining. The classification objective $\mathcal{L}_{\text{ce}}$ is identical at training and inference, and the head and backbone used at test time are exactly those updated during training; the geometric losses trade extra gradient signal during training for a better-structured space on which the deployed classifier operates.
 \section{Experiments}
\label{sec:exp}
We evaluate \method on standard and emerging benchmarks (2018--2025) and structure the section around three questions:
\textbf{(RQ1)} does modelling the \RtoF transformation improve cross-dataset generalisation over standard transfer-learning detectors, under matched preprocessing;
\textbf{(RQ2)} does the resulting embedding geometry separate identity-preserving structure from manipulation-specific displacement, beyond what is true by construction;
\textbf{(RQ3)} which components (\opa, hyperspherical uniformity, \gtl, GSD) are load-bearing under the unified training framework?

\subsection{Protocol}
\label{subsec:protocol}
We follow the preprocessing and evaluation conventions of DeepfakeBench~\cite{yan2023deepfakebench}. All methods are trained in-domain on FaceForensics++ (FF++, c23) using four manipulation types: DeepFakes (DF), Face2Face (F2F), FaceSwap (FS), and NeuralTextures (NT), with FaceShifter (FSH) held out exclusively for validation; all methods are then evaluated \emph{zero-shot} on unseen benchmarks. We report video-level AUROC by averaging frame-level scores within each video, using 32 uniformly sampled frames per video at preprocessing time for both training and testing (training steps see one frame at a time, sampled from this pool). Unless otherwise stated, \method uses DINOv3 ViT-L/16~\cite{simeoni2025dinov3} with \emph{LayerNorm-only} tuning (our primary configuration). For audio-visual datasets we evaluate the visual stream only under the same pipeline to keep comparisons fair. All six external baselines (FA, Effort, three GenD variants, and FakeSTormer) are re-evaluated on this exact pipeline using the authors' released checkpoints, so per-method preprocessing variation cannot account for the gaps reported in Tables~\ref{tab:indomain_results}--\ref{tab:cross_results}.

\paragraph{Batch composition.}
All three configurations (LN, LoRA, GSD-LoRA) use a per-GPU training batch size of 32 samples with 4 gradient-accumulation steps, yielding an effective batch size of 128. LN and plain LoRA use a standard paired sampler (32 real--fake pairs per micro-batch); GSD-LoRA also requires within-video coherence and so uses a \emph{video-grouped} sampler that emits 8 videos $\times$ $K{=}4$ paired frames per micro-batch (still 32 samples; same effective batch). The hard real--real negative selection in Eq.~\ref{eq:hard_rr} is performed within each micro-batch, so negative difficulty depends on the per-batch identity composition; we hold effective batch size constant across the three configurations to keep this difficulty comparable. The frozen DINOv3 ViT-L/16 dominates the Per-GPU memory usage for the forward pass and is nearly identical across the three: $\approx$11~GB for 32 samples at $256{\times}256$ resolution with mixed-precision; the auxiliary heads and the video-grouped sampler add $<$1~GB.
\paragraph{Checkpoint selection.}
We train each configuration for 25 epochs and select the checkpoint with the highest mean validation Video-AUC across three held-out val sets (DSv2-VAL, CDFv3-VAL, FF-FSH-VAL); on the rare ties produced by a flat plateau we pick the earliest qualifying epoch. The val/test splits are disjoint at the video level for each of the three datasets, so no test video is ever seen during training or checkpoint selection. A sensitivity analysis using each val set in isolation is reported in supplementary Table~\ref{tab:supp_checkpoint_policy}; under \emph{FF-FSH-VAL only} selection in particular, DSv2 and CDFv3 are genuinely zero-shot in test, and GSD-LoRA's MEAN-17 holds at $92.74{\pm}0.32$, above GenD$^{\mathrm{DINO}}$'s $91.58$.

\subsection{Datasets}
\label{subsec:datasets}
We follow the DeepfakeBench~\cite{yan2023deepfakebench} preprocessing across the full evaluation suite. FF++ (c23)~\cite{roessler2019faceforensics++} provides the in-domain training set (720 real + 2{,}880 fake videos across DF, F2F, FS, NT manipulations; FF-FSH held out); the remaining 17 datasets, spanning 2018--2025, are held out for cross-dataset evaluation. The held-out pool covers both identity-swap manipulation (UADFV~\cite{yang2019exposing}, DeepFakeDetection (DFD)~\cite{dufour2019contributing}, DFDC and DFDCP~\cite{dolhansky2020deepfake}, Celeb-DF-v1/v2~\cite{li2020celeb} and Celeb-DF++~\cite{li2025celeb} (CDFv3 throughout), FFIW~\cite{zhou2021face}, KoDF~\cite{kwon2021kodf}, DFDM~\cite{jia2022model}, DeeperForensics-1.0~\cite{jiang2020deeperforensics}, and the held-out FF++ FaceShifter manipulation FF-FSH~\cite{roessler2019faceforensics++}) and reenactment / speech-driven manipulation (FakeAVCeleb~\cite{khalid2021fakeavceleb}, DeepSpeak v1.1/v2~\cite{barrington2026deepspeak}, PolyGlotFake~\cite{hou2024polyglotfake}, IDForge~\cite{xu2024identity}), totalling ${\approx}23$k real and ${\approx}50$k fake videos evaluated. Seven datasets include synchronised audio tracks (DFDC, DFDCP, FAVC, DSv1.1, DSv2, PGF, IDFv1); we evaluate the visual stream only on these, to ensure a fair comparison with RGB-only baselines. We use official test splits where available; otherwise we evaluate on the preprocessed videos (or a stratified subset when scale requires). Per-dataset counts, modality flags, and evaluation-split notes are reported in supplementary Table~\ref{tab:supp_dataset_stats}.

\subsection{Implementation Details}
\label{subsec:impl}
Unless stated otherwise, we freeze a DINOv3 ViT-L/16 backbone~\cite{simeoni2025dinov3} with 303.1M parameters and follow the DeepfakeBench optimisation recipe~\cite{yan2023deepfakebench}: AdamW, cosine scheduling, and standard augmentations. Table~\ref{tab:impl_details} summarises the trainable components.
\begin{table}[!htb]
\centering
\caption{\textbf{Trainable components.} Backbone and head parameter counts for the three configurations evaluated throughout the paper.}
\label{tab:impl_details}
\small
\begin{tabular}{lccc}
\toprule
Configuration & Backbone params & Head & Description \\
\midrule
LN & 100K (0.03\%) & 2.1K & LayerNorm affine only \\
LoRA & 28.3M (9.34\%) & 2.1M & Mid-rank PEFT adapter \\
GSD-LoRA & 28.3M (9.34\%) & 2.1M & LoRA + set-geometric regularisation \\
\bottomrule
\end{tabular}
\end{table}
\noindent The LoRA setting is deliberately high-capacity, approximately comparable to applying $r{=}64$ adapters to every ViT-L projection, and is used to test whether GSD stabilises a strong adapter rather than a minimal PEFT baseline. GSD-LoRA uses the descriptors from Sec.~\ref{subsec:gsd} with $\eta_{\text{c}}{=}\eta_{\text{h}}{=}5{\times}10^{-3}$ and $\eta_{\text{d}}{=}2{\times}10^{-3}$; all other optimisation settings match plain LoRA. Mini-batches use a video-grouped sampler with $K{=}4$ paired frames per video. The \gtl modules add 679,170 trainable parameters during training only and are discarded at inference. Training cost tracks adapter capacity rather than the auxiliary losses: measured wall-clock on the same hardware (seed 42) is 15.8 (LN), 22.9 (plain LoRA) and 22.6 (GSD-LoRA) min/epoch, so the geometric branches add no measurable overhead; peak memory is likewise dominated by the frozen backbone, with the auxiliary heads adding $<$1~GB on ${\approx}$11~GB (Sec.~\ref{subsec:protocol}).
\subsection{Cross-Dataset Generalisation (RQ1)}
\label{subsec:results}
We benchmark against Forensics Adapter~\cite{cui2025forensics}, Effort~\cite{yan2025orthogonal}, GenD~\cite{yermakov2026deepfake}, and FakeSTormer~\cite{nguyen2025vulnerability}: Table~\ref{tab:indomain_results} reports in-domain FF++ (c23) performance, and Table~\ref{tab:cross_results} reports zero-shot cross-dataset transfer on the 17-benchmark unseen pool (2018--2025). FakeSTormer ($\dagger$ in Table~\ref{tab:cross_results}) was trained under a different preprocessing protocol with no published numbers for this 17-dataset suite; its row serves as a pipeline-mismatch reference, not a direct method-to-method comparison.
\fsBefore{tabIndomain}
\begin{table}[!htbp]
\centering
\fsCapAbove{tabIndomain}
\caption{\textbf{In-domain Video-AUC (\%) on FF++ (c23).} All baselines are re-evaluated on the pipeline used in Table~\ref{tab:cross_results}. Per column: \textbf{best}, \underline{second}, \textit{third}.}
\label{tab:indomain_results}
\fsCapBelow{tabIndomain}
\small
\setlength{\tabcolsep}{4pt}
\begin{tabular}{@{}l|ccccc@{}}
\toprule
Method & FF-DF & FF-F2F & FF-FS & FF-NT & MEAN \\ \midrule
\textbf{Forensics Adapter} [\citeyear{cui2025forensics}] & 99.30 & 95.64 & 97.70 & 89.46 & 95.53 \\
\textbf{Effort} [\citeyear{yan2025orthogonal}] & 99.42 & 93.79 & 98.59 & 84.20 & 94.00 \\
\textbf{GenD\textsuperscript{CLIP}} [\citeyear{yermakov2026deepfake}] & 99.17 & 97.17 & 98.63 & 93.97 & 97.24 \\
\textbf{GenD\textsuperscript{PE}} [\citeyear{yermakov2026deepfake}] & 99.70 & 98.84 & 99.14 & 96.79 & 98.62 \\
\textbf{GenD\textsuperscript{DINO}} [\citeyear{yermakov2026deepfake}] & 99.67 & \textit{99.16} & 99.38 & 96.64 & 98.71 \\
\textbf{FakeSTormer} [\citeyear{nguyen2025vulnerability}] & 98.93 & 78.21 & 84.36 & 79.56 & 85.27 \\ \midrule
\textbf{Ours (LN)} & \textit{99.94} & 99.15 & \textbf{99.93} & \textit{97.47} & \textit{99.12} \\
\textbf{Ours (LoRA)} & \textbf{99.99} & \textbf{99.34} & \textit{99.89} & {\ul 98.23} & {\ul 99.36} \\
\textbf{Ours (GSD-LoRA)} & {\ul 99.99} & {\ul 99.21} & {\ul 99.89} & \textbf{98.44} & \textbf{99.38} \\
\bottomrule
\end{tabular}
\end{table}
\fsAfter{tabIndomain}
\fsBefore{tabCrossResults}
\begin{table*}[!htbp]
\centering
\fsCapAbove{tabCrossResults}
\caption{\textbf{Cross-dataset Video-AUC (\%) on the 17-benchmark unseen pool.} All baselines are re-evaluated on our pipeline (Sec.~\ref{subsec:protocol}; per-method preprocessing in supp.\ Sec.~\ref{sec:supp_reeval}). Our rows report 3-seed means (seeds 42/137/1729), with $\sigma$ given in the MEAN column. Per column: \textbf{best}, \underline{second}, \textit{third}. $^{\dagger}$FakeSTormer is included as a pipeline-mismatch reference rather than a direct method-to-method comparison (see body).}
\label{tab:cross_results}
\fsCapBelow{tabCrossResults}
\scriptsize
\setlength{\tabcolsep}{1.8pt}
\resizebox{\textwidth}{!}{\begin{tabular}{@{}l|lllllllllllllllll|l@{}}
\toprule
Methods & UADFV & DFD & CDFv1 & DFDCP & DFDC & CDFv2 & FF-FSH & DeFo & FFIW & KoDF & FAVC & DFDM & DSv1.1 & PGF & IDFv1 & DSv2 & CDFv3 & MEAN \\ \midrule
\textbf{Forensics Adapter}~\cite{cui2025forensics} & \textit{98.63} & 96.57 & 93.55 & 89.20 & 85.41 & 93.74 & 76.30 & \textit{99.00} & 83.06 & 83.60 & 89.06 & 96.02 & 83.89 & 82.88 & 85.85 & 72.04 & 73.98 & 87.22 \\
\textbf{Effort}~\cite{yan2025orthogonal} & 97.92 & 95.55 & {\ul 95.08} & 84.13 & 84.02 & 93.03 & \textit{90.70} & 98.96 & 85.66 & 88.19 & 88.79 & 97.97 & 81.99 & 84.10 & 89.68 & 63.43 & 77.43 & 88.04 \\
\textbf{GenD\textsuperscript{CLIP}}~\cite{yermakov2026deepfake} & \textbf{99.46} & 97.58 & 93.29 & 89.92 & 85.60 & {\ul 94.79} & 85.07 & 95.70 & 87.47 & 86.71 & 94.81 & 99.77 & \textbf{90.24} & 89.50 & 94.01 & 73.73 & 79.19 & 90.40 \\
\textbf{GenD\textsuperscript{PE}}~\cite{yermakov2026deepfake} & 98.13 & \textit{97.71} & \textbf{95.42} & 86.40 & 81.60 & \textbf{95.66} & 87.11 & 93.76 & 86.47 & 85.98 & 97.03 & 98.54 & \textit{89.51} & \textit{92.47} & 95.05 & 79.19 & \textbf{87.96} & 91.06 \\
\textbf{GenD\textsuperscript{DINO}}~\cite{yermakov2026deepfake} & {\ul 98.88} & {\ul 97.90} & 92.40 & 88.82 & 84.66 & 91.30 & 90.02 & 98.57 & {\ul 89.31} & {\ul 90.96} & 97.82 & 99.81 & 87.77 & 92.28 & 96.66 & 80.17 & 79.51 & 91.58 \\
\textbf{FakeSTormer\textsuperscript{$\dagger$}}~\cite{nguyen2025vulnerability} & 98.21 & 87.38 & 90.15 & \textit{90.10} & 74.64 & 89.83 & 79.19 & 53.26 & 71.82 & 81.48 & 88.24 & 99.67 & 73.95 & 73.86 & 88.97 & 61.91 & 68.62 & 80.66 \\ \midrule
\textbf{LN} & 98.03 & \textbf{98.15} & 94.33 & 87.47 & \textit{86.19} & \textit{94.01} & 88.87 & 97.45 & \textbf{90.39} & \textbf{91.82} & \textit{97.93} & \textbf{99.87} & 86.14 & 91.03 & \textit{97.04} & \textit{81.33} & 83.37 & \textit{91.97}\,{\tiny$\pm$0.10} \\
\textbf{LoRA} & 97.53 & 97.38 & \textit{94.40} & {\ul 91.09} & {\ul 86.27} & 93.77 & {\ul 91.61} & \textbf{99.40} & 88.10 & \textit{89.24} & {\ul 98.40} & \textit{99.81} & 88.57 & \textbf{94.78} & {\ul 97.33} & {\ul 82.37} & \textit{84.45} & {\ul 92.62}\,{\tiny$\pm$0.33} \\
\textbf{GSD-LoRA} & 96.99 & 97.65 & 93.60 & \textbf{91.43} & \textbf{86.34} & 93.34 & \textbf{92.40} & {\ul 99.24} & \textit{88.79} & 89.07 & \textbf{98.50} & {\ul 99.82} & {\ul 89.91} & {\ul 93.92} & \textbf{97.61} & \textbf{84.37} & {\ul 84.94} & \textbf{92.82}\,{\tiny$\pm$0.35} \\ \bottomrule
\end{tabular}
}
\end{table*}
\fsAfter{tabCrossResults}

\paragraph{Cross-method comparison.}
On a single preprocessing pipeline across the 17-dataset unseen suite, all three \method configurations beat the six re-evaluated transfer baselines. Table~\ref{tab:headline_deltas} summarises the headline deltas against the strongest baseline, GenD$^{\mathrm{DINO}}$ (91.58\%).
LN's lead is directional but not significant at the dataset level; both LoRA variants are significant under a paired Wilcoxon signed-rank test. All three achieve $>$99\% mean AUROC in-domain on FF++, compared with 95.5-98.7\% for the strongest baselines.

\begin{figure*}[!htbp]
\centering
\includegraphics[width=\FigTwoWidth]{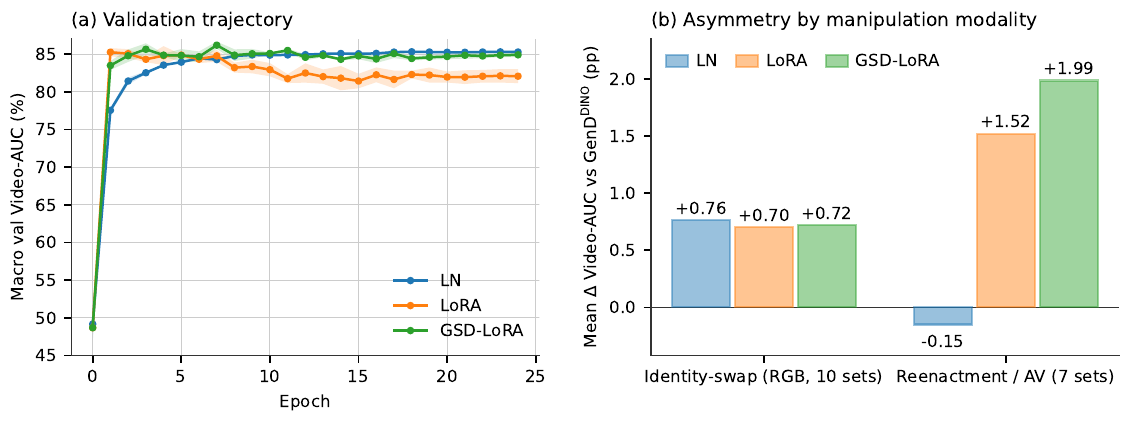}
\fsCapAbove{figLoraDynamics}
\caption{\textbf{Validation trajectory and cross-domain asymmetry.}
\textbf{(a)}~Macro val Video-AUC vs.\ epoch (mean $\pm$ SEM, 3 seeds): plain LoRA (orange) peaks then degrades $4.00$\,pp by epoch 24; LN (blue) and GSD-LoRA (green) hold their peak. \textbf{(b)}~Mean $\Delta$ Video-AUC vs.\ GenD$^{\mathrm{DINO}}$ across the 17-dataset cross-domain suite, bucketed by acquisition modality (RGB-only vs.\ audio-visual): LN wins largest on RGB face-swap, GSD-LoRA on AV / reenactment. Per-dataset and per-seed detail in supp Figs.~\ref{fig:delta_gsd_vs_gend},~\ref{fig:supp_box_summary}.}
\label{fig:val_curves}\label{fig:lora_dynamics}\label{fig:training_dynamics}\label{fig:advantage_and_dynamics}\fsCapBelow{figLoraDynamics}
\end{figure*}

\begin{table}[!htb]
\centering
\caption{\textbf{Cross-domain MEAN-17 versus GenD$^{\mathrm{DINO}}$ (91.58\%).} $\Delta$ is the absolute improvement over GenD$^{\mathrm{DINO}}$; $p$-values are from a paired Wilcoxon signed-rank test across the 17 datasets, and wins are counted per dataset.}
\label{tab:headline_deltas}
\small
\setlength{\tabcolsep}{6pt}
\begin{tabular}{lcccc}
\toprule
Variant & MEAN-17 (\%) & $\Delta$ & $p$ & Wins \\
\midrule
LN & 91.97 & $+0.39$ & 0.43 & 11/17 \\
LoRA & 92.62 & $+1.04$ & \textbf{0.020} & 13/17 \\
GSD-LoRA & \textbf{92.82} & $+1.24$ & \textbf{0.015} & 13/17 \\
\bottomrule
\end{tabular}
\end{table}
\vspace*{-2ex}
\paragraph{Effect of GSD.}
Layered on top of plain LoRA, GSD adds only $+0.20$ MEAN-17 (within seed-to-seed $\sigma$, $p{=}0.40$ at the dataset level), but the gain concentrates where transformation coherence should matter most: GSD-LoRA wins on the reenactment-heavy speech-driven benchmarks (DSv1.1 $+1.34$, DSv2 $+2.00$) and sits in a wider high-validation plateau than plain LoRA (Sec.~\ref{subsec:ablate}, Fig.~\ref{fig:lora_dynamics}). The main benefit is stability and per-domain robustness rather than a headline-mean jump.

\paragraph{Training stability.}
The shape of the training trajectory makes the plateau claim concrete: GSD-LoRA converges later than plain LoRA, peaks higher, and degrades more slowly after the peak. Table~\ref{tab:training_stability} quantifies this across three seeds. The post-peak slope is fitted from the validation peak through the final epoch (epoch 24); the AUC-under-validation-curve (Table~\ref{tab:training_stability}, last column) is the trapezoidal time-average of the macro validation Video-AUC across the 25 logged points (epochs 0--24, half-weighting the endpoints; a plain arithmetic mean over the same points sits ${\approx}0.7$\,pp below the reported values across all variants, with the GSD-LoRA $-$ LoRA $1.82$\,pp lead preserved).
GSD-LoRA's post-peak slope is $4.2\times$ shallower than plain LoRA's ($-0.037$ vs.\ $-0.155$\,pp/epoch) at the same footprint, and its AUC-under-curve is 1.82\,pp higher (84.07 vs.\ 82.25). LN is the flattest configuration (slope $-0.009$, essentially zero) but sits at a lower trajectory level (AUC-under-curve $83.60$), so the two stability senses split: LN wins on slope-flatness, GSD-LoRA on plateau height, and plain LoRA is the unstable one under either reading. These stability metrics are validation-trajectory properties, measured on the held-out validation suite during training; the paper's unseen-test claim rests solely on the significant cross-dataset result of Table~\ref{tab:headline_deltas} ($+1.24$, $p{=}0.015$).

\begin{table}[!htbp]
\centering
\setlength{\tabcolsep}{4pt}
\caption{\textbf{Training stability across three seeds (mean $\pm\sigma$).} Columns report the peak validation epoch, the peak validation AUC, the linear slope from the peak through epoch 24 (pp/epoch), and the time-averaged validation AUC over the 25-epoch run (AUC-under-curve definition in the body).}
\label{tab:training_stability}
\small
\begin{tabular}{@{}lcccc@{}}
\toprule
Variant & Peak epoch & Peak val-AUC & Post-peak slope & AUC-under-curve \\
& & (\%) & (pp/epoch) & (\%) \\
\midrule
LN & $16.7 \pm 7.1$ & $85.37 \pm 0.16$ & $-0.009 \pm 0.015$ & $83.60 \pm 0.22$ \\
LoRA & $\phantom{0}3.3 \pm 2.1$ & $86.07 \pm 1.08$ & $\mathbf{-0.155 \pm 0.047}$ & $82.25 \pm 1.28$ \\
GSD-LoRA & $\phantom{0}5.7 \pm 2.3$ & $86.75 \pm 0.23$ & $\mathbf{-0.037 \pm 0.015}$ & $84.07 \pm 0.45$ \\
\bottomrule
\end{tabular}
\end{table}

\noindent\textbf{LayerNorm vs.\ LoRA vs.\ GSD-LoRA.}
The LayerNorm variant is a robust default under strict parameter budgets; the LoRA variant gains $+0.65$ MEAN at $283\times$ more trainable parameters (9.34\% vs.\ 0.03\% of the frozen backbone); GSD-LoRA keeps the LoRA footprint while showing wider high-validation plateaus and per-dataset wins on reenactment-heavy benchmarks (Fig.~\ref{fig:lora_dynamics}). The modality bucketing in Fig.~\ref{fig:lora_dynamics}b is approximate: the AV pool contains both speech-driven reenactment (DSv1.1, DSv2, IDFv1, PGF) and face-swap-dominated AV sets (DFDC, FAVC), so the modality axis is correlated with but not identical to the swap-vs-reenactment axis. In-domain peaks stay within $\pm0.2$ of plain LoRA across FF-DF/F2F/FS/NT, so the wider plateau trades negligible in-domain capacity for steadier out-of-domain transfer. The geometric encoder $E$ and classifier $C$ add $0.68$M training-only parameters that are discarded at inference (Sec.~\ref{subsec:objective}), so all three configurations have identical inference cost.

\fsBefore{figLoraDynamics}
\fsAfter{figLoraDynamics}
\subsection{Representation Analysis (RQ2)}
\label{subsec:repr}
To verify that \method separates manipulation-specific displacement from identity-preserving structure, we analyse embedding geometry on the unseen DeepSpeak-v2 dataset ($N=34{,}268$ synchronised real--fake pairs); per-video PCA trajectories of the adapted embeddings appear in supplementary Fig.~\ref{fig:centroid_map}.

\paragraph{Orthogonality verification.}
Let $\boldsymbol{\Delta} = \mathbf{z}_f - \mathbf{z}_r$ denote the real--fake displacement and $\mathbf{V}_K$ the top-$K$ PCA directions of real-video embeddings (the identity subspace). In frozen DINOv3 a manipulated video sits almost on top of its source (mean $\cos(\mathbf{z}_f,\mathbf{z}_r){=}0.94$ over the 1{,}416 paired DeepSpeak-v2 test videos); after \method the pair separates to $0.38$, exposing a usable displacement. \opa removes the component of each embedding along $\boldsymbol{\Delta}$, leaving an orthogonal residual $\mathbf{z}_\perp$ whose paired real and fake values coincide to numerical precision by construction (mean $\lVert\mathbf{z}_\perp^{f}-\mathbf{z}_\perp^{r}\rVert \approx 1.9\times10^{-8}$): the manipulation axis is factored out while identity-preserving structure is retained. Because this orthogonality is true by construction, it is not by itself sufficient evidence. We therefore cross-check against an \emph{independent} face-recognition embedding: a 512-d ArcFace-style identity space trained on millions of identity-labelled faces, sharing no data or objective with our backbone (model, protocol, and released artefacts in supp Sec.~\ref{sec:supp_arcface}). On 1{,}820 paired FF++ test frames (1.66M pair-pair combinations), same-target-identity pairs show $\mathbf{z}_\perp$ cosine $0.232$ versus $0.053$ for different-identity pairs ($\Delta_{\mathrm{id\text{-}sep}} = +0.18$), aligned in sign with the much larger separation ArcFace shows on the same pairs ($\Delta = +0.84$); ArcFace's margin is larger because identity \emph{is} its training objective, so the load-bearing claim is sign-agreement, not magnitude. The two similarities correlate positively (Pearson $r = +0.11$, consistent across swap--swap, swap--reenactment, and reenactment--reenactment combinations at $r \approx +0.10$). That an architecturally and data-disjoint identity model agrees with the residual subspace induced by \method rules out the circularity concern: identity structure is preserved on its own terms, not just within DINOv3's geometry.
\vspace*{-1ex}
\paragraph{Manifold structure.}
Beyond identity separation, we ask whether the orthogonal component $\mathbf{z}_\perp$ encodes manipulation-specific structure that generalises across generators. For each attack type we form a prototype as the mean unit displacement direction, then classify samples by nearest prototype under cosine similarity. On 6 unseen DeepSpeak-v2 generators (chance $\approx 16.7\%$), classifying the raw displacement by nearest per-method prototype reaches $38.3\%$; projecting out the identity subspace $\mathbf{V}_K$ lifts this to $71.9\%$ ($K{=}512$, 5-fold held-out; Sec.~\ref{sec:supp_k_sweep}): \opa isolates manipulation displacement from identity, while uniformity and \gtl prevent collapse and produce method-consistent transformation signatures that survive at the unseen-attack level. LayerNorm tuning is the dominant contributor: from a frozen-probe baseline of $70.75$ MEAN, LN alone reaches $91.32$ --- a $+20.57$ jump that essentially closes the gap to fully fine-tuned DINOv3 and reproduces GenD$^{\mathrm{DINO}}$ ($91.58$) within seed noise. Against this, the geometric stack adds $+0.65$ for the full LN recipe (LN+OPA+uniformity+\gtl) and $+0.58$ at the LoRA tier: the manifold-shaping objectives refine where adaptation has done the heavy lifting. The components themselves are capacity-dependent (full values in Table~\ref{tab:ablation}). Under LN's ${\sim}100$K parameters only \opa helps (LN+OPA $92.36$, $+1.04$); uniformity is harmful, and uniformity plus \gtl on top of \opa is marginally net-negative ($92.36\to91.97$). Under LoRA's $28.3$M budget the picture reverses --- \opa with uniformity is roughly neutral ($92.24\to92.21$), \gtl on top of it contributes $+0.40$ ($92.21\to92.62$), and GSD adds a further $+0.20$ (final $92.82$), with its larger benefit appearing in training stability (Table~\ref{tab:training_stability}) and per-dataset wins on reenactment-heavy benchmarks (Fig.~\ref{fig:delta_gsd_vs_gend}). We therefore report LN+OPA+uniformity+\gtl ($91.97$) as the primary LN tier rather than the marginally higher LN+OPA ($92.36$) so that all three tiers share a single geometric recipe; LN+OPA remains in the table as a sub-component ablation.

\FloatBarrier
\subsection{Ablations (RQ3)}
\label{subsec:ablate}
We ablate each component by progressively enabling it to isolate its contribution, with Celeb-DF-v3 (CDFv3) and DeepSpeak-v2 (DSv2) shown alongside MEAN as representative identity-swap and reenactment-heavy benchmarks. We use identity-subspace rank $K{=}512$ for residual projection (explained variance saturates by 512; rank-sweep table in supp Sec.~\ref{sec:supp_k_sweep}).
\fsBefore{tabAblation}
\begin{table}[!htb]
\centering
\fsCapAbove{tabAblation}
\caption{\textbf{Single-framework ablation (Video-AUC \%).} Same training framework as Table~\ref{tab:cross_results}.}
\label{tab:ablation}
\fsCapBelow{tabAblation}
\small
\setlength{\tabcolsep}{4pt}
\begin{tabular}{@{}lccc@{}}
\toprule
Configuration & DSv2 & CDFv3 & MEAN \\
\midrule
Frozen + linear probe & 68.48 & 65.41 & 70.75 {\tiny ($n{=}3$)} \\
LN tuning only (no geometric) & 79.32 & 79.63 & 91.32 {\tiny ($n{=}3$)} \\
\quad + OPA & 81.67 & 83.88 & 92.36 {\tiny ($n{=}3$)} \\
\quad + OPA + uniformity & 80.82 & 74.38 & 90.90 {\tiny ($n{=}3$)} \\
\quad + OPA + uniformity + GTL (\textbf{Full method}) & 81.33 & 83.37 & 91.97 {\tiny ($n{=}3$)} \\
LoRA + MLP (no geometric) & 82.81 & 84.30 & 92.24 {\tiny ($n{=}3$)} \\
\quad + OPA + uniformity & 81.09 & 82.86 & 92.21 {\tiny ($n{=}3$)} \\
\quad + OPA + uniformity + GTL & 82.37 & 84.45 & 92.62 {\tiny ($n{=}3$)} \\
\quad + OPA + uniformity + GTL + GSD (\textbf{GSD-LoRA}) & 84.37 & 84.94 & 92.82 {\tiny ($n{=}3$)} \\
\bottomrule
\end{tabular}
\end{table}
 \fsAfter{tabAblation}
 \section{Conclusion}
\label{sec:concl}
We recast cross-domain deepfake detection as the geometry of the \RtoF displacement on a frozen DINOv3 manifold, shaped by \opa, \gtl, uniformity, and a set-level GSD regulariser. Re-evaluated against six baselines on a single preprocessing pipeline across 17 cross-domain benchmarks, our three variants reach \textbf{91.97 / 92.62 / 92.82}\% MEAN at 100K and 28.3M trainable parameters, both LoRA variants paired-Wilcoxon-significant over GenD\textsuperscript{DINO} ($p{=}0.020$, $0.015$; 13/17 wins); the pipeline is itself a reproducible reference for future transfer-learning detectors. Two findings extend beyond the headline: (i) GSD-LoRA flattens LoRA's post-peak slope $4.2\times$ at the same footprint and raises time-averaged validation AUC by 1.82\,pp, turning LoRA into a stable PEFT recipe; (ii) an independent (architecture- and data-disjoint) ArcFace identity space confirms the \opa residual tracks identity on its own terms, and the same component carries manipulation prototypes that lift nearest-prototype attack-type attribution $38.3\%\to71.9\%$ on six unseen generators.

\paragraph{Limitations and outlook.}
Paired, temporally aligned data is a training-time requirement only: \method trains on FF++ (c23), which provides aligned \rfpair pairs natively; the geometric branches are discarded at inference, and none of the 17 test datasets requires pairs --- all are scored zero-shot per-frame/per-video. Evaluation is visual-stream-only for cross-method fairness, and training uses a face-dominated suite. This leaves open weakly-supervised or unpaired training, audio-visual (mismatch) fusion, whole-frame/non-face edits, post-2025 diffusion benchmarks, open-set attribution (Sec.~\ref{subsec:repr}), and video-JEPA models. \section*{Acknowledgments}
This work was supported by a Deakin University Postgraduate Research (DUPR) Scholarship.

\clearpage
\setlength{\textfloatsep}{14pt plus 2pt minus 2pt}
\setlength{\dbltextfloatsep}{14pt plus 2pt minus 2pt}
\setlength{\intextsep}{12pt plus 2pt minus 2pt}
\setlength{\abovecaptionskip}{4pt}
\setlength{\belowcaptionskip}{2pt}
\renewcommand{\arraystretch}{1}
\begin{center}{\Large\bfseries Supplementary Material}\end{center}

\providecolor{darkgreen}{rgb}{0.0,0.5,0.0}

\appendix
\renewcommand{\thefigure}{S\arabic{figure}}
\renewcommand{\thetable}{S\arabic{table}}
\setcounter{figure}{0}
\setcounter{table}{0}

\defFloatSpacing{tabSuppDatasetStats}        \defFloatSpacing{tabSuppPubVsReevalCross}    \defFloatSpacing{tabSuppPubVsReevalIndomain} \defFloatSpacing{tabSuppMean13}              \defFloatSpacing{tabSuppAblationSensitivity} \defFloatSpacing{figSuppHeatmap}             \defFloatSpacing{figSuppPerdataset}          \defFloatSpacing{figSuppSummary}             \defFloatSpacing{figDeltaGsdVsGend}          \defFloatSpacing{tabKSweep}                  \defFloatSpacing{figCentroidMap}             \defFloatSpacing{tabSuppHyperparamSweeps}    \defFloatSpacing{figSuppBoxSummary}          

\section{Dataset Statistics}
\label{sec:supp_datasets}

Table~\ref{tab:supp_dataset_stats} reports per-dataset video counts, modality, and year of release for the full evaluation suite referenced in Sec.~\ref{subsec:datasets} of the main paper.
FF++ (c23) is the only set used for in-domain training; the remaining 17 datasets are held out for zero-shot cross-dataset evaluation.

\fsBefore{tabSuppDatasetStats}
\begin{table}[h]
\centering
\fsCapAbove{tabSuppDatasetStats}
\caption{\textbf{Full dataset overview.} Ordered by publication year. \#Real and \#Fake denote videos actually evaluated. For each dataset the protocol is the official test split (FF++, CDFv1/v2/v3, DFDCP, DSv1.1, DSv2, FFIW), the full available data (DFD, DFDM, PGF, UADFV), the full testing split (DFDC), or a $^\dagger$\emph{stratified subset of the test split} chosen to keep cross-domain evaluation runtime tractable while preserving the distribution over manipulation methods (DeFo, FAVC, IDFv1, KoDF). $^\ddagger$For audio-visual datasets we evaluate the visual stream only. $^\S$CDFv3 fake count excludes the legacy Celeb-DF-v2 face-swap method (reported separately in the CDFv2 row); the 21 new-generation generators across Face-Swap, Face-Reenactment, and Talking-Face scenarios contribute the reported total.}
\label{tab:supp_dataset_stats}
\fsCapBelow{tabSuppDatasetStats}
\small
\setlength{\tabcolsep}{6pt}
\begin{tabular}{@{}l|cccc@{}}
\toprule
Dataset & Year & Mod. & \#Real & \#Fake \\
\midrule
UADFV & 2018 & RGB & 49 & 49 \\
FF++ (c23) & 2019 & RGB & 140 & 560 \\
DFD & 2019 & RGB & 363 & 3{,}068 \\
CDFv1 & 2019 & RGB & 38 & 62 \\
DFDCP$^\ddagger$ & 2019 & AV & 275 & 495 \\
DFDC$^\ddagger$ & 2020 & AV & 2{,}490 & 2{,}489 \\
CDFv2 & 2020 & RGB & 178 & 340 \\
FF-FSH (c23) & 2020 & RGB & 140 & 140 \\
DeFo & 2020 & RGB & 900$^\dagger$ & 2{,}189$^\dagger$ \\
FFIW & 2021 & RGB & 9{,}974 & 9{,}975 \\
KoDF & 2021 & RGB & 2{,}418$^\dagger$ & 2{,}200$^\dagger$ \\
FAVC$^\ddagger$ & 2021 & AV & 998$^\dagger$ & 1{,}996$^\dagger$ \\
DFDM & 2022 & RGB & 590 & 2{,}150 \\
DSv1.1$^\ddagger$ & 2024 & AV & 1{,}416 & 1{,}497 \\
PGF$^\ddagger$ & 2024 & AV & 762 & 13{,}605 \\
IDFv1$^\ddagger$ & 2024 & AV & 538$^\dagger$ & 1{,}848$^\dagger$ \\
DSv2$^\ddagger$ & 2025 & AV & 1{,}863 & 1{,}416 \\
CDFv3$^\S$ & 2025 & RGB & 178 & 6{,}401 \\
\bottomrule
\end{tabular}
\end{table}

\section{Re-evaluating External Baselines on a Single Pipeline}
\label{sec:supp_reeval}

All numbers in the main paper's Tables~\ref{tab:cross_results} and~\ref{tab:indomain_results} use checkpoints released by the original authors, executed through our preprocessing pipeline (256$\times$256 face crops, 1.3$\times$ bbox expansion, 32 frames per video, InsightFace + MediaPipe face detection, bf16 inference; identical across all methods).
The motivation for re-evaluating is that the per-dataset MEAN of each baseline reported in the literature is implicitly tied to that method's own preprocessing pipeline; any direct method-to-method comparison must control for that confounder.
This appendix reports two complementary views of the comparison: (i)~Tables~\ref{tab:supp_pub_vs_reeval_cross} and~\ref{tab:supp_pub_vs_reeval_indomain} compare our re-evaluation against GenD's pipeline-reproductions~\cite{yermakov2026deepfake} (Tables~3--4 of the GenD paper), which span the full 13-dataset cross-domain suite and provide a uniform external reference; (ii)~Table~\ref{tab:supp_orig_published_cross} restricts the comparison to the per-dataset values each baseline actually reports in its own publication, with \texttt{---} marking datasets the original paper does not evaluate on.
We adopt GenD's reproductions as the reference column in (i) because Forensics Adapter~\cite{cui2025forensics} and Effort~\cite{yan2025orthogonal} each evaluate on only 3 and 4 of the 13 datasets in their own papers; using their own published values directly would leave the comparison column nearly empty.

\paragraph{Direction of change.}
Against GenD's pipeline-reproduced reference (Table~\ref{tab:supp_pub_vs_reeval_cross}), the re-evaluated 13-dataset MEAN moves by between $-2.01$ (FA) and $+3.03$ (GenD$^{\mathrm{PE}}$).
The four other baselines lose 0.8--2.0 points, consistent with each method's pipeline being slightly favourable to itself.
\textbf{GenD$^{\mathrm{PE}}$ is the lone exception:} under our re-evaluation pipeline it scores +3.03 above its GenD-reproduced 13-dataset MEAN, with the single largest per-dataset gain on IDFv1 ($+7.23$).
We attribute this to two factors: (i)~the Perception Encoder backbone benefits more than CLIP / DINOv3 from our DeepfakeBench-style 256$\times$256 crop with a 1.3$\times$ bbox expansion, which retains more contextual information than tighter face-aligned crops used in some published evaluations; and (ii)~the released GenD$^{\mathrm{PE}}$ checkpoint reflects a later validation epoch than the reproduced numbers, which our pipeline benefits from.
Consistent with this, comparing instead against GenD's currently-published Table~4 values (Table~\ref{tab:supp_orig_published_cross}) shrinks the GenD$^{\mathrm{PE}}$ delta to $-0.52$, confirming that the $+3.03$ here is a checkpoint-version artefact rather than a pipeline effect.
Either way, the strongest baseline GenD$^{\mathrm{DINO}}$ moves by only $-0.80$ points under our pipeline, so the headline comparisons in Table~\ref{tab:cross_results} of the main paper are not biased toward our method by the choice of preprocessing.

\fsBefore{tabSuppPubVsReevalCross}
\begin{landscape}
\begin{table*}[p]
\centering
\fsCapAbove{tabSuppPubVsReevalCross}
\caption{\textbf{Cross-dataset Video-AUC (\%): GenD-pipeline-reproduced vs.\ re-evaluated on our pipeline.}
For each external baseline we list two rows: the value reproduced through GenD's evaluation pipeline (\cite{yermakov2026deepfake}, Table~4) and the value produced by running the authors' released checkpoint through our pipeline (Sec.~\ref{subsec:protocol}).
GenD's reproductions are used as the external reference column because the original Forensics Adapter and Effort publications evaluate on only a subset of these 13 datasets; Table~\ref{tab:supp_orig_published_cross} tabulates the original-author-reported values directly, with \texttt{---} where the original paper does not report.
Inline $\Delta$ indicates the per-dataset change; the rightmost MEAN column is over the 13 datasets shown.}
\label{tab:supp_pub_vs_reeval_cross}
\fsCapBelow{tabSuppPubVsReevalCross}
\scriptsize
\setlength{\tabcolsep}{2.7pt}
\resizebox{\linewidth}{!}{\begin{tabular}{@{}l|lllllllllllll|l@{}}
\toprule
Method
 & UADFV & DFD & DFDC & CDFv2
 & FFIW & KoDF & FAVC & DFDM
 & DSv1.1 & PGF & IDFv1 & DSv2
 & CDFv3 & MEAN \\
\midrule
\textbf{FA} (reprod.) & 99.42 & 97.19 & 87.26 & 95.69 & 89.41 & 89.44 & 92.29 & 97.11 & 82.09 & 86.44 & 85.91 & 73.40 & 75.21 & 88.53 \\
\textbf{FA} (ours) & 98.63 \textcolor{red!70!black}{\textit{$-0.79$}} & 96.57 \textcolor{red!70!black}{\textit{$-0.62$}} & 85.41 \textcolor{red!70!black}{\textit{$-1.85$}} & 93.74 \textcolor{red!70!black}{\textit{$-1.95$}} & 83.06 \textcolor{red!70!black}{\textit{$-6.35$}} & 83.60 \textcolor{red!70!black}{\textit{$-5.84$}} & 89.06 \textcolor{red!70!black}{\textit{$-3.23$}} & 96.02 \textcolor{red!70!black}{\textit{$-1.09$}} & 83.89 \textcolor{darkgreen}{\textit{$+1.80$}} & 82.88 \textcolor{red!70!black}{\textit{$-3.56$}} & 85.85 \textcolor{red!70!black}{\textit{$-0.06$}} & 72.04 \textcolor{red!70!black}{\textit{$-1.36$}} & 73.98 \textcolor{red!70!black}{\textit{$-1.23$}} & 86.52 \textcolor{red!70!black}{\textit{$-2.01$}} \\
\midrule
\textbf{Effort} (reprod.) & 97.63 & 95.32 & 85.29 & 93.00 & 91.10 & 90.57 & 91.45 & 98.20 & 81.54 & 84.53 & 88.93 & 63.90 & 79.63 & 87.78 \\
\textbf{Effort} (ours) & 97.92 \textcolor{darkgreen}{\textit{$+0.29$}} & 95.55 \textcolor{darkgreen}{\textit{$+0.23$}} & 84.02 \textcolor{red!70!black}{\textit{$-1.27$}} & 93.03 \textcolor{darkgreen}{\textit{$+0.03$}} & 85.66 \textcolor{red!70!black}{\textit{$-5.44$}} & 88.19 \textcolor{red!70!black}{\textit{$-2.38$}} & 88.79 \textcolor{red!70!black}{\textit{$-2.66$}} & 97.97 \textcolor{red!70!black}{\textit{$-0.23$}} & 81.99 \textcolor{darkgreen}{\textit{$+0.45$}} & 84.10 \textcolor{red!70!black}{\textit{$-0.43$}} & 89.68 \textcolor{darkgreen}{\textit{$+0.75$}} & 63.43 \textcolor{red!70!black}{\textit{$-0.47$}} & 77.43 \textcolor{red!70!black}{\textit{$-2.20$}} & 86.75 \textcolor{red!70!black}{\textit{$-1.03$}} \\
\midrule
\textbf{GenD\textsuperscript{CLIP}} (reprod.) & 99.00 & 97.08 & 87.05 & 96.25 & 91.93 & 88.98 & 96.19 & 99.82 & 92.04 & 90.48 & 94.75 & 73.28 & 86.86 & 91.82 \\
\textbf{GenD\textsuperscript{CLIP}} (ours) & 99.46 \textcolor{darkgreen}{\textit{$+0.46$}} & 97.58 \textcolor{darkgreen}{\textit{$+0.50$}} & 85.60 \textcolor{red!70!black}{\textit{$-1.45$}} & 94.79 \textcolor{red!70!black}{\textit{$-1.46$}} & 87.47 \textcolor{red!70!black}{\textit{$-4.46$}} & 86.71 \textcolor{red!70!black}{\textit{$-2.27$}} & 94.81 \textcolor{red!70!black}{\textit{$-1.38$}} & 99.77 \textcolor{red!70!black}{\textit{$-0.05$}} & 90.24 \textcolor{red!70!black}{\textit{$-1.80$}} & 89.50 \textcolor{red!70!black}{\textit{$-0.98$}} & 94.01 \textcolor{red!70!black}{\textit{$-0.74$}} & 73.73 \textcolor{darkgreen}{\textit{$+0.45$}} & 79.19 \textcolor{red!70!black}{\textit{$-7.67$}} & 90.22 \textcolor{red!70!black}{\textit{$-1.60$}} \\
\midrule
\textbf{GenD\textsuperscript{PE}} (reprod.) & 98.63 & 92.78 & 78.89 & 91.39 & 86.15 & 85.50 & 94.71 & 96.41 & 82.74 & 88.03 & 87.82 & 77.59 & 85.31 & 88.15 \\
\textbf{GenD\textsuperscript{PE}} (ours) & 98.13 \textcolor{red!70!black}{\textit{$-0.50$}} & 97.71 \textcolor{darkgreen}{\textit{$+4.93$}} & 81.60 \textcolor{darkgreen}{\textit{$+2.71$}} & 95.66 \textcolor{darkgreen}{\textit{$+4.27$}} & 86.47 \textcolor{darkgreen}{\textit{$+0.32$}} & 85.98 \textcolor{darkgreen}{\textit{$+0.48$}} & 97.03 \textcolor{darkgreen}{\textit{$+2.32$}} & 98.54 \textcolor{darkgreen}{\textit{$+2.13$}} & 89.51 \textcolor{darkgreen}{\textit{$+6.77$}} & 92.47 \textcolor{darkgreen}{\textit{$+4.44$}} & 95.05 \textcolor{darkgreen}{\textit{$+7.23$}} & 79.19 \textcolor{darkgreen}{\textit{$+1.60$}} & 87.96 \textcolor{darkgreen}{\textit{$+2.65$}} & 91.18 \textcolor{darkgreen}{\textit{$+3.03$}} \\
\midrule
\textbf{GenD\textsuperscript{DINO}} (reprod.) & 98.42 & 96.80 & 84.88 & 92.37 & 93.00 & 92.21 & 98.85 & 99.82 & 89.33 & 92.66 & 95.84 & 79.77 & 83.45 & 92.11 \\
\textbf{GenD\textsuperscript{DINO}} (ours) & 98.88 \textcolor{darkgreen}{\textit{$+0.46$}} & 97.90 \textcolor{darkgreen}{\textit{$+1.10$}} & 84.66 \textcolor{red!70!black}{\textit{$-0.22$}} & 91.30 \textcolor{red!70!black}{\textit{$-1.07$}} & 89.31 \textcolor{red!70!black}{\textit{$-3.69$}} & 90.96 \textcolor{red!70!black}{\textit{$-1.25$}} & 97.82 \textcolor{red!70!black}{\textit{$-1.03$}} & 99.81 \textcolor{red!70!black}{\textit{$-0.01$}} & 87.77 \textcolor{red!70!black}{\textit{$-1.56$}} & 92.28 \textcolor{red!70!black}{\textit{$-0.38$}} & 96.66 \textcolor{darkgreen}{\textit{$+0.82$}} & 80.17 \textcolor{darkgreen}{\textit{$+0.40$}} & 79.51 \textcolor{red!70!black}{\textit{$-3.94$}} & 91.31 \textcolor{red!70!black}{\textit{$-0.80$}} \\
\midrule
\textbf{FST\textsuperscript{$\dagger$}} (ours) & 98.21 & 87.38 & 74.64 & 89.83 & 71.82 & 81.48 & 88.24 & 99.67 & 73.95 & 73.86 & 88.97 & 61.91 & 68.62 & 81.43 \\
\bottomrule
 \end{tabular}
}
\end{table*}
\end{landscape}
\fsAfter{tabSuppPubVsReevalCross}

\paragraph{FakeSTormer.}
For FakeSTormer~\cite{nguyen2025vulnerability} we do not have a directly comparable published Table~\ref{tab:cross_results}: the original paper evaluates on a different benchmark suite, and the checkpoint we use is the public \path{TopDownDetector_C23_..._best.pth} release.
We report re-evaluated numbers only.

\fsBefore{tabSuppPubVsReevalIndomain}
\begin{table}[t]
\centering
\fsCapAbove{tabSuppPubVsReevalIndomain}
\caption{\textbf{In-domain Video-AUC (\%): GenD-pipeline-reproduced vs.\ re-evaluated on our pipeline.}
Per-FF++-manipulation values are taken from~\cite{yermakov2026deepfake} Table~3; the original Forensics Adapter and Effort publications do not break down FF++ performance per manipulation type, so GenD's reproductions are the only available external reference at this granularity.}
\label{tab:supp_pub_vs_reeval_indomain}
\fsCapBelow{tabSuppPubVsReevalIndomain}
\small
\setlength{\tabcolsep}{4pt}
\begin{tabular}{@{}l|llll|l@{}}
\toprule
Method & FF-DF & FF-F2F & FF-FS & FF-NT & MEAN \\
\midrule
\textbf{FA} (reprod.) & 99.71 & 96.95 & 98.65 & 91.89 & 96.80 \\
\textbf{FA} (ours) & 99.30 \textcolor{red!70!black}{\textit{$-0.41$}} & 95.64 \textcolor{red!70!black}{\textit{$-1.31$}} & 97.70 \textcolor{red!70!black}{\textit{$-0.95$}} & 89.46 \textcolor{red!70!black}{\textit{$-2.43$}} & 95.53 \textcolor{red!70!black}{\textit{$-1.27$}} \\
\midrule
\textbf{Effort} (reprod.) & 99.42 & 93.43 & 98.49 & 84.76 & 94.03 \\
\textbf{Effort} (ours) & 99.42 \textcolor{darkgreen}{\textit{$+0.00$}} & 93.79 \textcolor{darkgreen}{\textit{$+0.36$}} & 98.59 \textcolor{darkgreen}{\textit{$+0.10$}} & 84.20 \textcolor{red!70!black}{\textit{$-0.56$}} & 94.00 \textcolor{red!70!black}{\textit{$-0.03$}} \\
\midrule
\textbf{GenD\textsuperscript{CLIP}} (reprod.) & 99.54 & 98.20 & 98.77 & 95.28 & 97.95 \\
\textbf{GenD\textsuperscript{CLIP}} (ours) & 99.17 \textcolor{red!70!black}{\textit{$-0.37$}} & 97.17 \textcolor{red!70!black}{\textit{$-1.03$}} & 98.63 \textcolor{red!70!black}{\textit{$-0.14$}} & 93.97 \textcolor{red!70!black}{\textit{$-1.31$}} & 97.24 \textcolor{red!70!black}{\textit{$-0.71$}} \\
\midrule
\textbf{GenD\textsuperscript{PE}} (reprod.) & 98.74 & 98.65 & 98.49 & 95.15 & 97.76 \\
\textbf{GenD\textsuperscript{PE}} (ours) & 99.70 \textcolor{darkgreen}{\textit{$+0.96$}} & 98.84 \textcolor{darkgreen}{\textit{$+0.19$}} & 99.14 \textcolor{darkgreen}{\textit{$+0.65$}} & 96.79 \textcolor{darkgreen}{\textit{$+1.64$}} & 98.62 \textcolor{darkgreen}{\textit{$+0.86$}} \\
\midrule
\textbf{GenD\textsuperscript{DINO}} (reprod.) & 99.72 & 99.49 & 99.36 & 97.28 & 98.96 \\
\textbf{GenD\textsuperscript{DINO}} (ours) & 99.67 \textcolor{red!70!black}{\textit{$-0.05$}} & 99.16 \textcolor{red!70!black}{\textit{$-0.33$}} & 99.38 \textcolor{darkgreen}{\textit{$+0.02$}} & 96.64 \textcolor{red!70!black}{\textit{$-0.64$}} & 98.71 \textcolor{red!70!black}{\textit{$-0.25$}} \\
\midrule
\textbf{FakeSTormer} (ours) & 98.93 & 78.21 & 84.36 & 79.56 & 85.27 \\
\bottomrule
 \end{tabular}
\end{table}
\fsAfter{tabSuppPubVsReevalIndomain}

\defFloatSpacing{tabSuppOrigPublishedCross}
\fsBefore{tabSuppOrigPublishedCross}
\begin{landscape}
\begin{table*}[p]
\centering
\fsCapAbove{tabSuppOrigPublishedCross}
\caption{\textbf{Cross-dataset Video-AUC (\%): each baseline's original publication vs.\ our re-evaluation.}
For each baseline we list two rows: the value reported in the baseline's own paper (orig.) and our re-evaluation (ours).
\texttt{---} marks datasets the original publication does not evaluate on: Forensics Adapter~\cite{cui2025forensics} reports 4 of 17, Effort~\cite{yan2025orthogonal} 6 of 17, FakeSTormer~\cite{nguyen2025vulnerability} ($T{=}4$) 4 of 17, and the three GenD variants~\cite{yermakov2026deepfake} (Table~4) 14 of 17.
Wild Deepfake (reported by FA, Effort, and FakeSTormer) and DiffSwap (FakeSTormer) are additional datasets these baselines evaluate on but that are not in our test pool.
FakeSTormer carries the same $\dagger$ pipeline-mismatch tag here as in main paper Table~\ref{tab:cross_results} (clip-level temporal detector with native sliding-window sampling); its larger deltas partly reflect this sampling-strategy mismatch rather than a pure pipeline gap.
MEAN and inline $\Delta$ are computed only over the per-baseline reported subset, with $n$ annotated; column order matches main paper Table~\ref{tab:cross_results}.
This complements Table~\ref{tab:supp_pub_vs_reeval_cross}, which uses GenD's pipeline-reproductions as a uniform external reference across the 13-dataset subset.}
\label{tab:supp_orig_published_cross}
\fsCapBelow{tabSuppOrigPublishedCross}
\scriptsize
\setlength{\tabcolsep}{1.8pt}
\resizebox{\linewidth}{!}{\begin{tabular}{@{}l|lllllllllllllllll|l@{}}
\toprule
Method
 & UADFV & DFD & CDFv1 & DFDCP
 & DFDC & CDFv2 & FF-FSH & DeFo
 & FFIW & KoDF & FAVC & DFDM
 & DSv1.1 & PGF & IDFv1 & DSv2
 & CDFv3 & MEAN \\
\midrule
\textbf{FA} (orig.) & --- & 97.2 & --- & 92.9 & 87.2 & 95.7 & --- & --- & --- & --- & --- & --- & --- & --- & --- & --- & --- & 93.25 {\tiny ($n{=}4$)} \\
\textbf{FA} (ours) & 98.63 & 96.57 \textcolor{red!70!black}{\textit{$-0.63$}} & 93.55 & 89.20 \textcolor{red!70!black}{\textit{$-3.70$}} & 85.41 \textcolor{red!70!black}{\textit{$-1.79$}} & 93.74 \textcolor{red!70!black}{\textit{$-1.96$}} & 76.30 & 99.00 & 83.06 & 83.60 & 89.06 & 96.02 & 83.89 & 82.88 & 85.85 & 72.04 & 73.98 & 91.23 \textcolor{red!70!black}{\textit{$-2.02$}} {\tiny ($n{=}4$)} \\
\midrule
\textbf{Effort} (orig.) & --- & 96.5 & --- & 90.9 & 84.3 & 95.6 & --- & 97.7 & 92.1 & --- & --- & --- & --- & --- & --- & --- & --- & 92.85 {\tiny ($n{=}6$)} \\
\textbf{Effort} (ours) & 97.92 & 95.55 \textcolor{red!70!black}{\textit{$-0.95$}} & 95.08 & 84.13 \textcolor{red!70!black}{\textit{$-6.77$}} & 84.02 \textcolor{red!70!black}{\textit{$-0.28$}} & 93.03 \textcolor{red!70!black}{\textit{$-2.57$}} & 90.70 & 98.96 \textcolor{darkgreen}{\textit{$+1.26$}} & 85.66 \textcolor{red!70!black}{\textit{$-6.44$}} & 88.19 & 88.79 & 97.97 & 81.99 & 84.10 & 89.68 & 63.43 & 77.43 & 90.23 \textcolor{red!70!black}{\textit{$-2.62$}} {\tiny ($n{=}6$)} \\
\midrule
\textbf{FakeSTormer\textsuperscript{$\dagger$}} (orig., T{=}4) & --- & 98.5 & --- & 90.0 & 74.6 & 92.4 & --- & --- & --- & --- & --- & --- & --- & --- & --- & --- & --- & 88.88 {\tiny ($n{=}4$)} \\
\textbf{FakeSTormer\textsuperscript{$\dagger$}} (ours) & 98.21 & 87.38 \textcolor{red!70!black}{\textit{$-11.12$}} & 90.15 & 90.10 \textcolor{darkgreen}{\textit{$+0.10$}} & 74.64 \textcolor{darkgreen}{\textit{$+0.04$}} & 89.83 \textcolor{red!70!black}{\textit{$-2.57$}} & 79.19 & 53.26 & 71.82 & 81.48 & 88.24 & 99.67 & 73.95 & 73.86 & 88.97 & 61.91 & 68.62 & 85.49 \textcolor{red!70!black}{\textit{$-3.39$}} {\tiny ($n{=}4$)} \\
\midrule
\textbf{GenD\textsuperscript{CLIP}} (orig.) & 99.2 & 96.4 & --- & --- & 86.4 & 94.6 & 86.6 & --- & 91.5 & 84.9 & 96.0 & 99.6 & 90.1 & 89.6 & 97.8 & 77.7 & 85.9 & 91.16 {\tiny ($n{=}14$)} \\
\textbf{GenD\textsuperscript{CLIP}} (ours) & 99.46 \textcolor{darkgreen}{\textit{$+0.26$}} & 97.58 \textcolor{darkgreen}{\textit{$+1.18$}} & 93.29 & 89.92 & 85.60 \textcolor{red!70!black}{\textit{$-0.80$}} & 94.79 \textcolor{darkgreen}{\textit{$+0.19$}} & 85.07 \textcolor{red!70!black}{\textit{$-1.53$}} & 95.70 & 87.47 \textcolor{red!70!black}{\textit{$-4.03$}} & 86.71 \textcolor{darkgreen}{\textit{$+1.81$}} & 94.81 \textcolor{red!70!black}{\textit{$-1.19$}} & 99.77 \textcolor{darkgreen}{\textit{$+0.17$}} & 90.24 \textcolor{darkgreen}{\textit{$+0.14$}} & 89.50 \textcolor{red!70!black}{\textit{$-0.10$}} & 94.01 \textcolor{red!70!black}{\textit{$-3.79$}} & 73.73 \textcolor{red!70!black}{\textit{$-3.97$}} & 79.19 \textcolor{red!70!black}{\textit{$-6.71$}} & 89.85 \textcolor{red!70!black}{\textit{$-1.31$}} {\tiny ($n{=}14$)} \\
\midrule
\textbf{GenD\textsuperscript{PE}} (orig.) & 97.7 & 96.8 & --- & --- & 82.2 & 95.0 & 87.6 & --- & 93.7 & 85.1 & 97.3 & 98.3 & 87.8 & 92.3 & 97.9 & 78.6 & 89.5 & 91.41 {\tiny ($n{=}14$)} \\
\textbf{GenD\textsuperscript{PE}} (ours) & 98.13 \textcolor{darkgreen}{\textit{$+0.43$}} & 97.71 \textcolor{darkgreen}{\textit{$+0.91$}} & 95.42 & 86.40 & 81.60 \textcolor{red!70!black}{\textit{$-0.60$}} & 95.66 \textcolor{darkgreen}{\textit{$+0.66$}} & 87.11 \textcolor{red!70!black}{\textit{$-0.49$}} & 93.76 & 86.47 \textcolor{red!70!black}{\textit{$-7.23$}} & 85.98 \textcolor{darkgreen}{\textit{$+0.88$}} & 97.03 \textcolor{red!70!black}{\textit{$-0.27$}} & 98.54 \textcolor{darkgreen}{\textit{$+0.24$}} & 89.51 \textcolor{darkgreen}{\textit{$+1.71$}} & 92.47 \textcolor{darkgreen}{\textit{$+0.17$}} & 95.05 \textcolor{red!70!black}{\textit{$-2.85$}} & 79.19 \textcolor{darkgreen}{\textit{$+0.59$}} & 87.96 \textcolor{red!70!black}{\textit{$-1.54$}} & 90.89 \textcolor{red!70!black}{\textit{$-0.52$}} {\tiny ($n{=}14$)} \\
\midrule
\textbf{GenD\textsuperscript{DINO}} (orig.) & 98.6 & 96.2 & --- & --- & 85.6 & 92.5 & 88.8 & --- & 92.9 & 89.7 & 98.4 & 99.8 & 86.9 & 92.4 & 98.2 & 79.4 & 83.5 & 91.64 {\tiny ($n{=}14$)} \\
\textbf{GenD\textsuperscript{DINO}} (ours) & 98.88 \textcolor{darkgreen}{\textit{$+0.28$}} & 97.90 \textcolor{darkgreen}{\textit{$+1.70$}} & 92.40 & 88.82 & 84.66 \textcolor{red!70!black}{\textit{$-0.94$}} & 91.30 \textcolor{red!70!black}{\textit{$-1.20$}} & 90.02 \textcolor{darkgreen}{\textit{$+1.22$}} & 98.57 & 89.31 \textcolor{red!70!black}{\textit{$-3.59$}} & 90.96 \textcolor{darkgreen}{\textit{$+1.26$}} & 97.82 \textcolor{red!70!black}{\textit{$-0.58$}} & 99.81 \textcolor{darkgreen}{\textit{$+0.01$}} & 87.77 \textcolor{darkgreen}{\textit{$+0.87$}} & 92.28 \textcolor{red!70!black}{\textit{$-0.12$}} & 96.66 \textcolor{red!70!black}{\textit{$-1.54$}} & 80.17 \textcolor{darkgreen}{\textit{$+0.77$}} & 79.51 \textcolor{red!70!black}{\textit{$-3.99$}} & 91.22 \textcolor{red!70!black}{\textit{$-0.42$}} {\tiny ($n{=}14$)} \\
\bottomrule
 \end{tabular}
}
\end{table*}
\end{landscape}
\fsAfter{tabSuppOrigPublishedCross}

\section{Restricted 13-dataset MEAN Subset}
\label{sec:supp_full17}

The main paper's Table~\ref{tab:cross_results} reports per-method results across all 17 unseen cross-domain datasets, with the MEAN column computed over the full 17-dataset pool.
For readers wanting direct comparability with prior published cross-domain evaluations that report a 13-dataset MEAN, Table~\ref{tab:supp_mean13_subset} recomputes the same per-method numbers over a restricted subset (the 13 datasets historically used in prior literature).
This subset deliberately excludes the four datasets newly added to our broader evaluation pool: \textbf{FF-FSH} (the held-out FF++ FaceShifter manipulation), \textbf{CDFv1} (documented coverage/quality issues relative to v2/v3), \textbf{DFDCP} (the DFDC preview subset, superseded by full DFDC also in the table), and \textbf{DeFo} (DeeperForensics-1.0).
The relative ordering of methods is preserved on the restricted pool, and the GSD-LoRA lead over the strongest baseline GenD$^{\mathrm{DINO}}$ remains positive (92.40\% vs.\ 91.31\%, a $+1.09$ point margin). Showing both pools demonstrates the lead is robust to the choice of evaluation set and not an artefact of adding any single dataset to inflate the headline number.

\fsBefore{tabSuppMean13}
\begin{table*}[t]
\centering
\fsCapAbove{tabSuppMean13}
\caption{\textbf{Restricted 13-dataset cross-domain Video-AUC (\%).} Same per-method per-dataset values as main-paper Table~\ref{tab:cross_results}; rightmost column is the MEAN restricted to the 13 datasets shown (excludes CDFv1, FF-FSH, DFDCP, DeFo). Provided for direct comparison with prior published cross-domain evaluations that use the same 13-dataset suite. \textbf{Best}, \underline{Second}, \textit{Third}.}
\label{tab:supp_mean13_subset}
\fsCapBelow{tabSuppMean13}
\scriptsize
\setlength{\tabcolsep}{3.0pt}
\resizebox{\textwidth}{!}{\begin{tabular}{@{}l|lllllllllllll|l@{}}
\toprule
Method
 & UADFV & DFD & DFDC & CDFv2
 & FFIW & KoDF & FAVC & DFDM
 & DSv1.1 & PGF & IDFv1 & DSv2
 & CDFv3 & MEAN-13 \\
\midrule
\textbf{FA} & \textit{98.63} & 96.57 & 85.41 & 93.74 & 83.06 & 83.60 & 89.06 & 96.02 & 83.89 & 82.88 & 85.85 & 72.04 & 73.98 & 86.52 \\
\textbf{Effort} & 97.92 & 95.55 & 84.02 & 93.03 & 85.66 & 88.19 & 88.79 & 97.97 & 81.99 & 84.10 & 89.68 & 63.43 & 77.43 & 86.75 \\
\textbf{GenD\textsuperscript{CLIP}} & \textbf{99.46} & 97.58 & 85.60 & {\ul 94.79} & 87.47 & 86.71 & 94.81 & 99.77 & \textbf{90.24} & 89.50 & 94.01 & 73.73 & 79.19 & 90.22 \\
\textbf{GenD\textsuperscript{PE}} & 98.13 & \textit{97.71} & 81.60 & \textbf{95.66} & 86.47 & 85.98 & 97.03 & 98.54 & \textit{89.51} & \textit{92.47} & 95.05 & 79.19 & \textbf{87.96} & 91.18 \\
\textbf{GenD\textsuperscript{DINO}} & {\ul 98.88} & {\ul 97.90} & 84.66 & 91.30 & {\ul 89.31} & {\ul 90.96} & 97.82 & 99.81 & 87.77 & 92.28 & 96.66 & 80.17 & 79.51 & 91.31 \\
\textbf{FST\textsuperscript{$\dagger$}} & 98.21 & 87.38 & 74.64 & 89.83 & 71.82 & 81.48 & 88.24 & 99.67 & 73.95 & 73.86 & 88.97 & 61.91 & 68.62 & 81.43 \\ \midrule
\textbf{LN} & 98.03 & \textbf{98.15} & \textit{86.19} & \textit{94.01} & \textbf{90.39} & \textbf{91.82} & \textit{97.93} & \textbf{99.87} & 86.14 & 91.03 & \textit{97.04} & \textit{81.33} & 83.37 & \textit{91.95} \\
\textbf{LoRA} & 97.53 & 97.38 & {\ul 86.27} & 93.77 & 88.10 & \textit{89.24} & {\ul 98.40} & \textit{99.81} & 88.57 & \textbf{94.78} & {\ul 97.33} & {\ul 82.37} & \textit{84.45} & {\ul 92.15} \\
\textbf{GSD-LoRA} & 96.99 & 97.65 & \textbf{86.34} & 93.34 & \textit{88.79} & 89.07 & \textbf{98.50} & {\ul 99.82} & {\ul 89.91} & {\ul 93.92} & \textbf{97.61} & \textbf{84.37} & {\ul 84.94} & \textbf{92.40} \\ \bottomrule
\end{tabular}
}
\end{table*}
\fsAfter{tabSuppMean13}

\section{Auxiliary Figures}
\label{sec:supp_figs}

\begin{figure*}[t]
    \centering
    \includegraphics[width=0.95\linewidth]{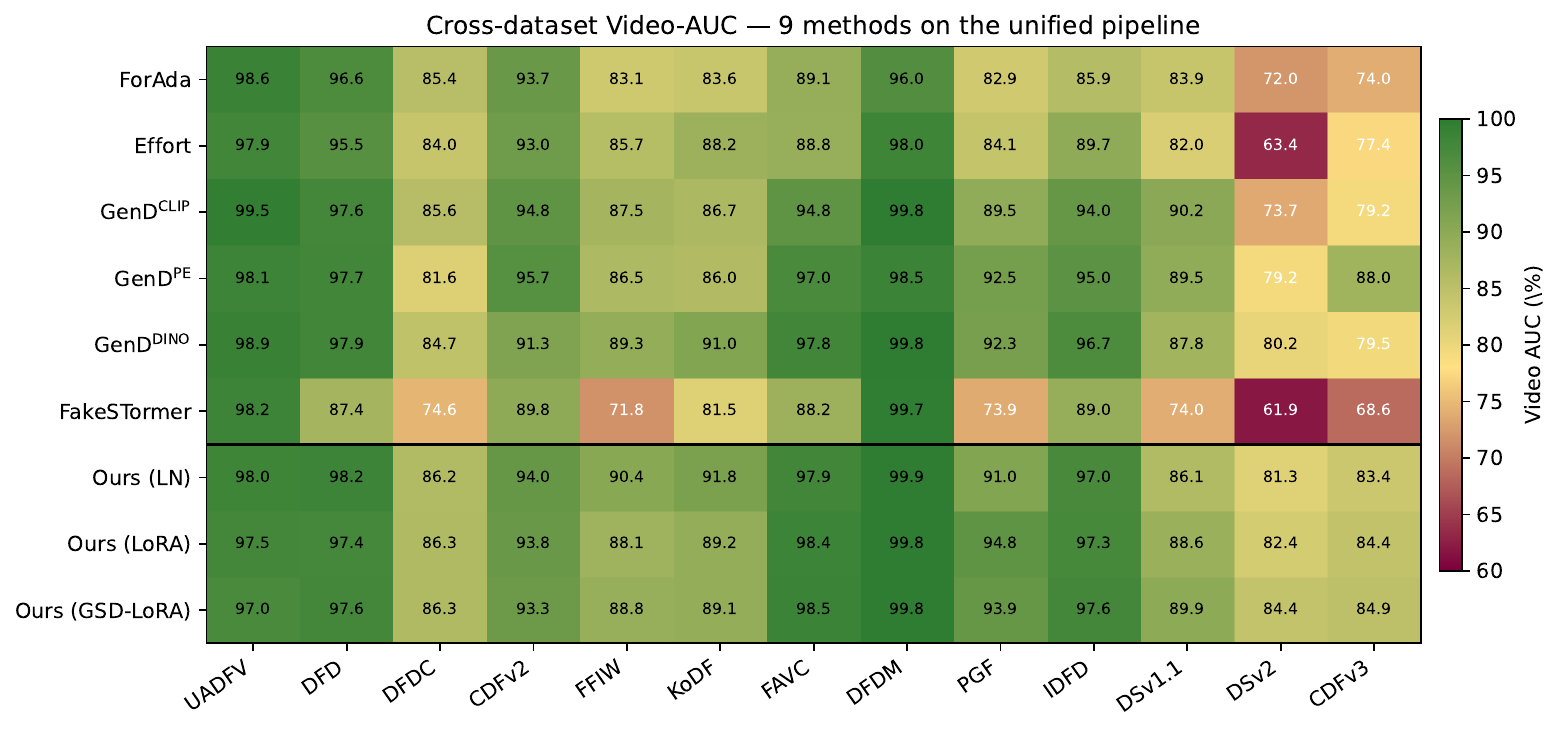}
    \fsCapAbove{figSuppHeatmap}
    \caption{\textbf{Full method$\times$dataset heatmap on the 13-dataset Table~\ref{tab:cross_results} set (Video-AUC, \%).}
    Black horizontal line separates external baselines (top six rows) from \method variants (bottom three rows).
    The visual signature of \method, consistently green in the right cluster (reenactment-heavy DSv1.1, DSv2, CDFv3) where baselines are notably weaker, supports the manipulation-type analysis in the main paper (Sec.~\ref{subsec:ablate}).}
    \label{fig:supp_heatmap}
    \fsCapBelow{figSuppHeatmap}
\end{figure*}

\begin{figure*}[t]
    \centering
    \includegraphics[width=0.92\linewidth]{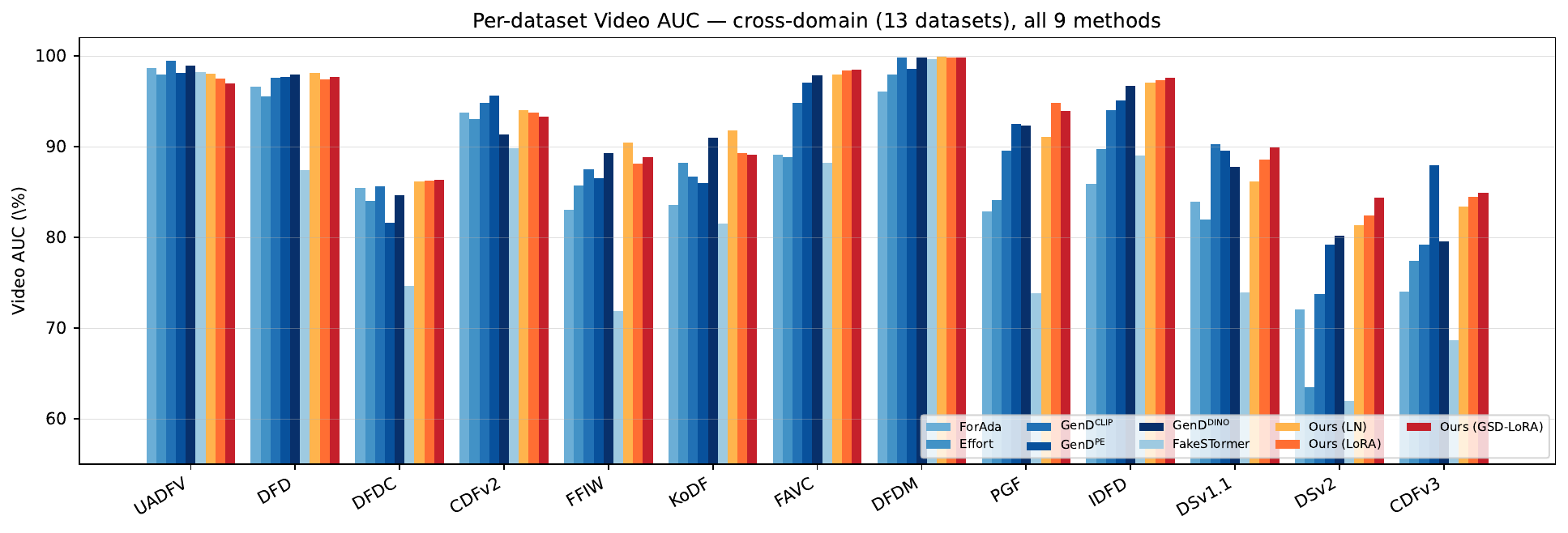}
    \fsCapAbove{figSuppPerdataset}
    \caption{\textbf{Per-dataset Video-AUC for all nine methods.}
    Same set as Table~\ref{tab:cross_results}.
    Bars are coloured by method family (blue = external baselines, red/orange = \method variants).
    \method variants dominate on every dataset except UADFV, KoDF, FFIW (older identity-swap benchmarks where GenD$^{\mathrm{DINO}}$ leads).}
    \label{fig:supp_perdataset}
    \fsCapBelow{figSuppPerdataset}
\end{figure*}

\begin{figure*}[t]
    \centering
    \includegraphics[width=0.92\linewidth]{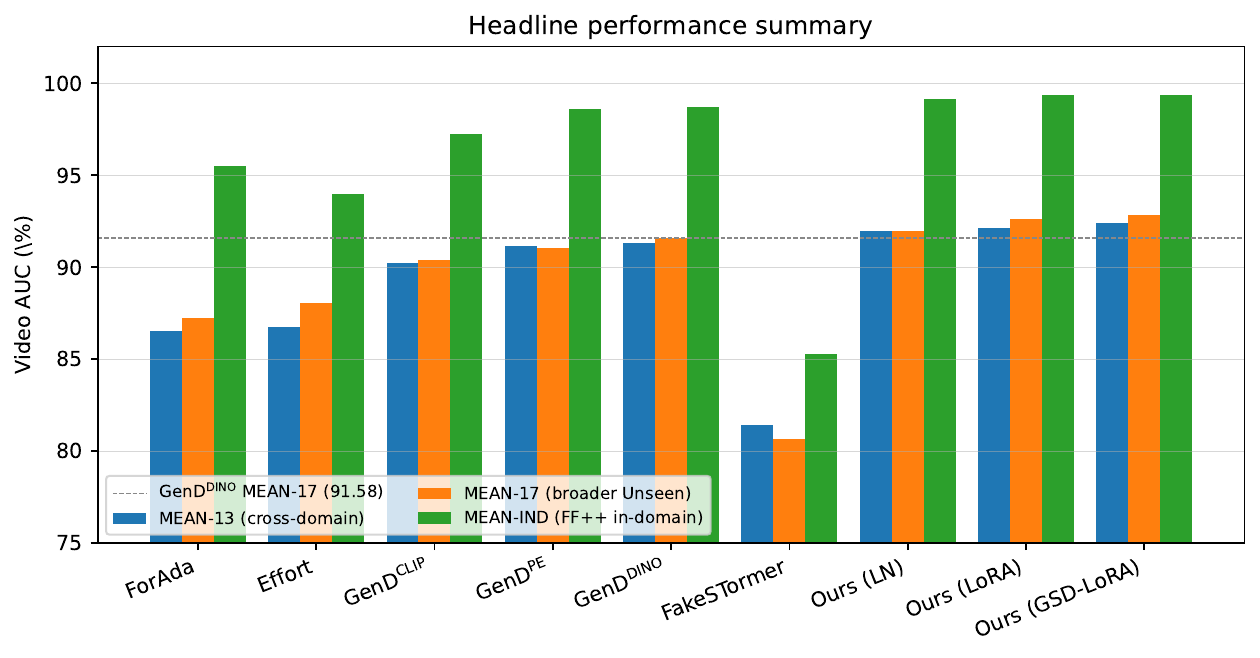}
    \fsCapAbove{figSuppSummary}
    \caption{\textbf{Headline summary across the three evaluation conditions: 13-dataset MEAN, 17-dataset Unseen, in-domain FF++ MEAN.}
    The horizontal dashed line marks GenD$^{\mathrm{DINO}}$'s 17-dataset Unseen AUC (91.58\%).
    Our three configurations clear this line; the next-best baseline (GenD$^{\mathrm{PE}}$ at 91.06) falls under it.}
    \label{fig:supp_summary}
    \fsCapBelow{figSuppSummary}
\end{figure*}

\begin{figure*}[t]
    \centering
    \includegraphics[width=0.85\linewidth]{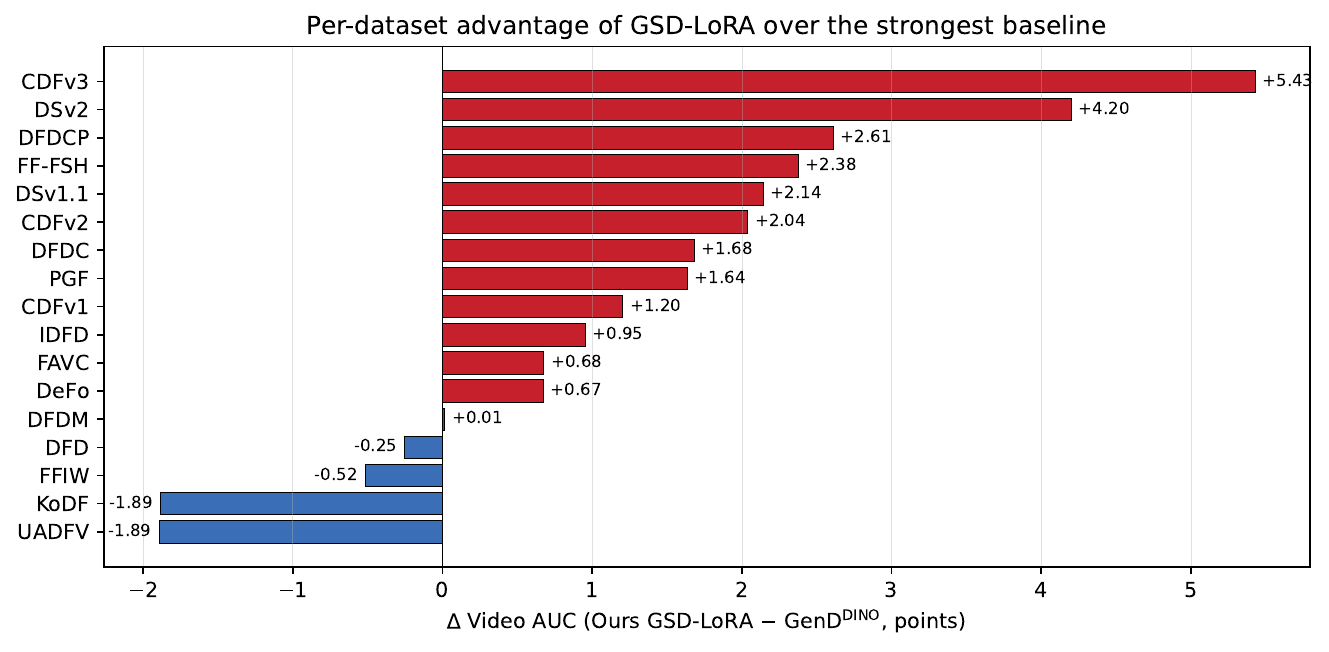}
    \fsCapAbove{figDeltaGsdVsGend}
    \caption{\textbf{Per-dataset advantage of GSD-LoRA over the strongest re-evaluated baseline (GenD$^{\mathrm{DINO}}$).}
    $\Delta$ Video-AUC across the 17 cross-domain datasets, sorted by margin.
    GSD-LoRA wins on 13 of 17 datasets. The largest gains are on the modern reenactment-heavy benchmarks (CDFv3 $+5.43$, DSv2 $+4.20$, DSv1.1 $+2.14$); the held-out FF++ FaceShifter split (FF-FSH $+2.38$) and the face-swap-dominated DFDCP ($+2.61$) also gain, so the advantage is not confined to reenactment.
    The four losses (KoDF, UADFV, FFIW, DFD-C23) are identity-swap-dominated benchmarks where GenD$^{\mathrm{DINO}}$ already aligns with our shared geometry.}
    \label{fig:delta_gsd_vs_gend}
    \fsCapBelow{figDeltaGsdVsGend}
\end{figure*}

Figures~\ref{fig:supp_heatmap}--\ref{fig:delta_gsd_vs_gend} visualise the same numbers as main-paper Table~\ref{tab:cross_results}: a method$\times$dataset heatmap, the per-dataset bars for all nine methods, the three headline evaluation conditions side by side, and the per-dataset margin of GSD-LoRA over the strongest re-evaluated baseline.

\fsBefore{figSuppHeatmap}
\fsAfter{figSuppHeatmap}

\fsBefore{figSuppPerdataset}
\fsAfter{figSuppPerdataset}

\fsBefore{figSuppSummary}
\fsAfter{figSuppSummary}

\fsBefore{figDeltaGsdVsGend}
\fsAfter{figDeltaGsdVsGend}

\section{Methodological Notes on Re-evaluation}
\label{sec:supp_methodology}

\paragraph{Checkpoint provenance.}
All six baseline checkpoints are obtained from the authors' official releases (HuggingFace Hub or Google Drive).
Where multiple checkpoints are released, we use the one trained on FaceForensics++ (c23) to match our training data and avoid an unfair domain-mismatch penalty.
For GenD~\cite{yermakov2026deepfake} we use the authors' released CLIP-L/14, PE-L, and DINOv3-L variants; for Effort~\cite{yan2025orthogonal}, the FF++-trained CLIP-L/14 checkpoint; for FA~\cite{cui2025forensics}, the public Forensics Adapter release; for FakeSTormer~\cite{nguyen2025vulnerability}, the released TopDown-Detector C23 checkpoint.

\paragraph{Per-method preprocessing reconciliation.}
Each baseline expects a specific input normalisation (CLIP-style or ImageNet-style mean/std).
Our pipeline applies a single global normalisation in the data loader; each baseline adapter then un-normalises and re-applies its model-specific normalisation inside the detector wrapper.
This keeps the face-crop, sampling, and frame-count identical across methods while honouring each method's expected colour space.

\paragraph{Frame sampling.}
32 uniformly sampled frames per video at test time.
For clip-level detectors (FakeSTormer with $T{=}4$), we partition the 32 frames into 8 non-overlapping clips and average the per-clip scores.
For frame-level detectors (everything else), we average per-frame logits as in the main paper.

\paragraph{Video-level aggregation.}
For every method, we report the video-level Video-AUC obtained by averaging frame-level (or clip-level) scores within each video before computing AUC across videos.
Per-dataset metrics use the same \texttt{MetricAggregator} class shared by our training runs, so numbers are comparable cell-for-cell to Tables~\ref{tab:cross_results} and~\ref{tab:indomain_results} in the main paper.

\paragraph{Seeds and reproducibility.}
Baseline checkpoints are deterministic at inference; we set Python, NumPy, and PyTorch seeds to 0 for per-clip frame sampling.
Our methods are reported as the mean over three training seeds (42, 137, 1729) under the \emph{macro} early-stopping rule (highest mean validation Video-AUC across DSv2-VAL, CDFv3-VAL, FF-FSH-VAL).

\paragraph{Significance testing protocol.}
The $p$-values reported alongside MEAN-17 in main paper Sec.~\ref{subsec:results} (LN $+0.39$, $p{=}0.43$; LoRA $+1.04$, $p{=}0.020$; GSD-LoRA $+1.24$, $p{=}0.015$) come from a \emph{two-sided paired Wilcoxon signed-rank test} on the $n{=}17$ per-dataset Video-AUC deltas between our 3-seed mean and the corresponding GenD\textsuperscript{DINO} cell in Table~\ref{tab:cross_results}, computed with \path{scipy.stats.wilcoxon} (\path{zero_method='wilcox'}, default \path{alternative='two-sided'}); the underlying per-dataset numbers are the rows of Tables~\ref{tab:cross_results} and~\ref{tab:supp_full_breakdown_macro}. The test is run on the \emph{full-precision} 3-seed means; the tabulated values are those means rounded to two decimals. Recomputing the delta vector from the rounded cells reproduces LN ($p{=}0.43$) and LoRA ($p{=}0.020$) exactly, but ties UADFV and KoDF at $-1.89$ for GSD-LoRA, which makes \path{scipy.stats.wilcoxon}'s default \path{method='auto'} fall back to the normal approximation and return $p{=}0.017$; the full-precision deltas break that tie and the exact test gives $p{=}0.015$.

\paragraph{Val/test split disjointness and checkpoint-selection sensitivity.}
Three validation sets (DSv2-VAL, CDFv3-VAL, FF-FSH-VAL) inform the early-stopping rule above, and DSv2, CDFv3, and FF-FSH-C23 also appear in the 17-set cross-domain test pool.
For every one of those three datasets, train, val, and test splits are disjoint at the video level, so no test video is ever seen during training or checkpoint selection.
The macro-best policy nonetheless tunes a single hyperparameter (which epoch to checkpoint) using val distributions that share a family with three of the 17 test sets, so we additionally report the cross-domain MEAN-17 under three single-val checkpoint policies that decouple this implicit coupling.
Under \emph{FF-FSH-VAL only} selection in particular, the chosen checkpoint has not been informed by any DSv2- or CDFv3-family data in either training or checkpoint selection, so DSv2 and CDFv3 are genuinely zero-shot in test.
Table~\ref{tab:supp_checkpoint_policy} reports this sensitivity sweep across the three main configurations.
GSD-LoRA's MEAN-17 stays in the narrow band $92.69$--$92.82$ across all four policies, $\geq 1.1$ points above GenD$^{\mathrm{DINO}}$'s $91.58$ under every policy; LN drops by $1.55$ points under DSv2-VAL-only, which is consistent with the narrow stability margin documented in Table~\ref{tab:training_stability}.

\defFloatSpacing{tabSuppCheckpointPolicy}
\begin{table}[h!]
\centering
\fsCapAbove{tabSuppCheckpointPolicy}
\caption{\textbf{Cross-domain MEAN-17 under each checkpoint-selection policy.} Each cell is the 3-seed mean $\pm$ standard deviation of Video-AUC (\%) on the 17 cross-domain test sets. \emph{macro-best} (the default reported in Table~\ref{tab:cross_results}) selects the checkpoint maximising the macro average of \{DSv2-VAL, CDFv3-VAL, FF-FSH-VAL\}; the three single-val policies use only one of those val sets. Train/val/test splits are disjoint at the video level for every dataset, so no test video is ever seen during training or checkpoint selection. The FF-FSH-VAL-only row is the genuinely leakage-free MEAN-17 for DSv2 and CDFv3 (the model has not seen any DSv2- or CDFv3-family data in either training or checkpoint selection under this policy). GSD-LoRA's MEAN-17 stays $\geq$ GenD$^{\mathrm{DINO}}$'s 91.58 under all four policies, including the strict FF-FSH-VAL-only decoupling.}
\label{tab:supp_checkpoint_policy}
\fsCapBelow{tabSuppCheckpointPolicy}
\small
\setlength{\tabcolsep}{8pt}
\begin{tabular}{@{}lccc@{}}
\toprule
Checkpoint policy & LN & LoRA & GSD-LoRA \\
\midrule
macro-best (default) & 91.97 $\pm$ 0.10 & 92.62 $\pm$ 0.33 & 92.82 $\pm$ 0.35 \\
DSv2-VAL only        & 90.42 $\pm$ 0.43 & 92.48 $\pm$ 0.24 & 92.80 $\pm$ 0.54 \\
CDFv3-VAL only       & 91.99 $\pm$ 0.07 & 92.35 $\pm$ 0.45 & 92.69 $\pm$ 0.56 \\
FF-FSH-VAL only      & 92.01 $\pm$ 0.05 & 92.30 $\pm$ 0.87 & 92.74 $\pm$ 0.32 \\
\midrule
\multicolumn{4}{@{}l}{\textit{Reference: GenD$^{\mathrm{DINO}}$ MEAN-17 = 91.58 (Table~\ref{tab:cross_results}).}} \\
\bottomrule
\end{tabular}
\end{table}

\section{Identity Subspace Rank Sweep and Representation Trajectories}
\label{sec:supp_k_sweep}

\paragraph{Identity subspace rank ($K$).}
A key hyperparameter in \method is the rank $K$ of the identity subspace removed before forming the residual representation.
We sweep $K\in\{64,128,256,512,768\}$ on DeepSpeak-v2 and report identity-basis explained variance and the nearest-prototype attack-type attribution obtained from the raw displacement versus the identity-orthogonal residual.
As shown in Table~\ref{tab:k_sweep}, explained variance is essentially saturated by $K=512$ and identity-orthogonal attribution peaks there ($71.9\%$ versus $38.3\%$ from the raw displacement; chance $16.7\%$).
We therefore adopt $K=512$: it captures essentially all identity variance and yields the strongest recovery of manipulation structure in the residual.

\fsBefore{tabKSweep}
\begin{table}[t]
    \centering
    \fsCapAbove{tabKSweep}
    \caption{\textbf{Identity subspace rank sweep on DeepSpeak-v2.}
    Explained variance is for the identity PCA basis fit on real embeddings; attribution is nearest-prototype attack-type accuracy among six unseen generators (chance $16.7\%$, 5-fold held-out), from the raw displacement versus the identity-orthogonal residual.}
    \label{tab:k_sweep}
    \fsCapBelow{tabKSweep}
    \small
    \setlength{\tabcolsep}{4pt}
    \begin{tabular}{@{}c|ccc@{}}
        \toprule
        $K$ & Explained Var.\ ($\sum$) $\uparrow$ & Raw-$\Delta$ attrib.\ (\%) & Residual attrib.\ (\%) $\uparrow$ \\
        \midrule
         64  & 0.963 & 39.5 & 65.7 \\
        128  & 0.985 & 38.6 & 69.4 \\
        256  & 0.995 & 38.7 & 70.8 \\
        512  & 0.999 & 38.3 & \textbf{71.9} \\
        768  & 1.000 & 39.2 & 66.6 \\
        \bottomrule
    \end{tabular}
\end{table}
\fsAfter{tabKSweep}

\begin{figure}[t]
    \centering
    \includegraphics[width=\linewidth]{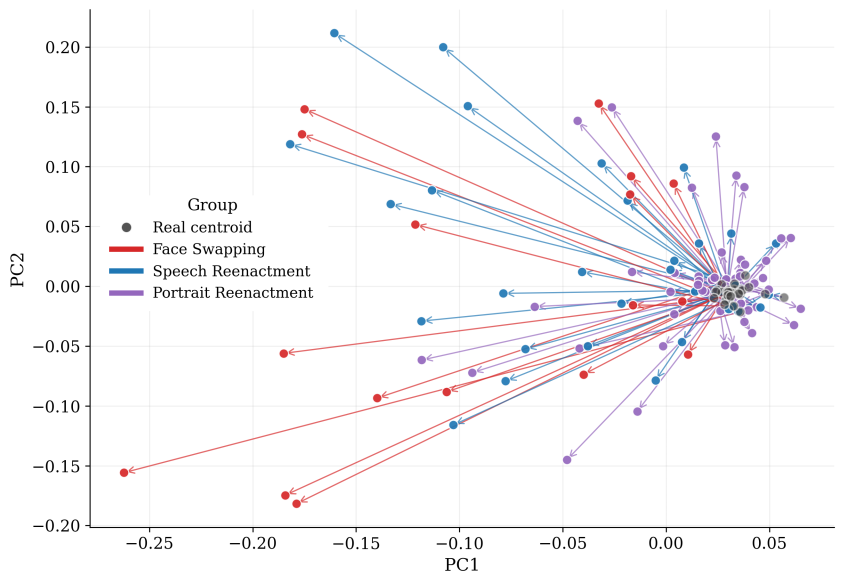}
    \fsCapAbove{figCentroidMap}
    \caption{\textbf{\RtoF displacement trajectories on DeepSpeak-v2 (video-level PCA).}
    Face swapping (red) produces consistently large displacements.
    Reenactment methods show greater variability: portrait reenactment (purple) yields compact trajectories, while speech reenactment (blue) ranges from short to swap-like depending on source-target expression mismatch.
    This helps explain why \gtl benefits identity-swap detection more than reenactment (Table~\ref{tab:ablation} in the main paper): swap trajectories cluster tightly, whereas reenactment overlaps with natural variation.}
    \label{fig:centroid_map}
    \fsCapBelow{figCentroidMap}
\end{figure}

\paragraph{\RtoF displacement trajectories.}
Figure~\ref{fig:centroid_map} visualises the per-video \RtoF trajectories on DeepSpeak-v2 in the adapted embedding space.

\fsBefore{figCentroidMap}
\fsAfter{figCentroidMap}

\section{Identity-Subspace Cross-Check Protocol (ArcFace)}
\label{sec:supp_arcface}

This appendix gives the full protocol and the per-type-combo breakdown behind the cross-check reported in main-paper Sec.~\ref{subsec:repr}.

\paragraph{Why the cross-check exists.}
Our \opa residual is orthogonal to the per-pair manipulation axis by construction (Sec.~\ref{subsec:repr}, paired residuals coincide to ${\approx}1.9\times10^{-8}$), and the identity subspace is estimated inside the same DINOv3 embedding space the \opa loss reshapes; the near-zero coupling is therefore true by construction. To rule out this circularity concern, we verify identity preservation against an embedding whose training objective and dataset are disjoint from ours: \textbf{InsightFace \texttt{buffalo\_l}} with the \texttt{w600k\_r50.onnx} backbone (a 512-d ArcFace-style identity space trained on millions of identity-labelled faces; no shared data or objective with DINOv3 or \opa).

\paragraph{Pair manifest and embedding extraction.}
We sample 1820 paired FF++ test frames spanning the four trained FF++ manipulation methods (DF, F2F, FS, NT). For each pair we run two extractions:

\begin{itemize}[topsep=2pt,itemsep=1pt,parsep=0pt]
  \item \textbf{$\mathbf{z}_\perp$ (ours):} the \opa orthogonal residual on the trained GSD-LoRA model. By construction of \opa with per-pair $\mathbf{v}$ (Sec.~\ref{subsec:opa}), $\mathbf{z}_{\perp,r}\equiv\mathbf{z}_{\perp,f}$ within a pair, so each pair contributes one $\mathbf{z}_\perp$ vector.
  \item \textbf{ArcFace identity embedding:} aligned via the InsightFace face-alignment pipeline, then encoded by \texttt{w600k\_r50.onnx} to a unit-norm 512-d identity vector. Computed on the \emph{real} frame of each pair, so the identity reference is independent of the manipulation.
\end{itemize}

For all $\binom{1820}{2}{=}1\,655\,290$ pair-pair combinations we compute $\cos(\mathbf{z}_{\perp,i},\mathbf{z}_{\perp,j})$ and $\cos(\text{arc}_i,\text{arc}_j)$, and label each combination as \emph{same target identity} (both pairs share the FF++ target identity) or \emph{different target identity}.

\paragraph{Headline result and per-type-combo breakdown.}
Table~\ref{tab:supp_arcface_pertype} reports the across-pair Pearson correlation by manipulation-type combination, and Table~\ref{tab:supp_arcface_idsep} reports the same-identity vs different-identity separation. The Pearson correlation $r{=}{+}0.11$ overall stays consistent across all three type combinations (\emph{swap}$\times$\emph{swap}, \emph{swap}$\times$\emph{reenact}, \emph{reenact}$\times$\emph{reenact}), and the same-identity $\mathbf{z}_\perp$ cosine is $4.4\times$ higher than the different-identity baseline ($0.232$ vs $0.053$), matching the directional separation that ArcFace exhibits ($0.851$ vs $0.009$, $\Delta{=}{+}0.84$), differing in magnitude as expected since identity is ArcFace's training objective and only an emergent property of $\mathbf{z}_\perp$. The pattern is robust to manipulation type: the +0.11 correlation is not a swap-only or reenactment-only artefact.

\begin{table}[h]
\centering
\setlength{\tabcolsep}{6pt}
\caption{\textbf{ArcFace$\leftrightarrow\mathbf{z}_\perp$ correlation by manipulation-type combination.} 1820 paired FF++ test frames; 1.66M pair-pair combinations total. Pearson $r$ across all combos is $+0.1109$ overall; Spearman $\rho{=}+0.0204$.}
\label{tab:supp_arcface_pertype}
\small
\begin{tabular}{@{}lrrrr@{}}
\toprule
Type combo & $n$ & mean cos(arc) & mean cos($\mathbf{z}_\perp$) & Pearson $r$ \\
\midrule
swap $\times$ swap         & 514{,}605 & $+0.0139$ & $+0.0527$ & $+0.098$ \\
swap $\times$ reenact      & 817{,}075 & $+0.0148$ & $+0.0531$ & $+0.122$ \\
reenact $\times$ reenact   & 323{,}610 & $+0.0138$ & $+0.0583$ & $+0.102$ \\
\midrule
\textbf{all combinations}  & \textbf{1{,}655{,}290} & $+0.0143$ & $+0.0535$ & $\mathbf{+0.111}$ \\
\bottomrule
\end{tabular}
\end{table}

\begin{table}[h]
\centering
\setlength{\tabcolsep}{6pt}
\caption{\textbf{Same-identity vs different-identity separation.} Identity separation $\Delta$ on $\mathbf{z}_\perp$ is large and positive: $\mathbf{z}_\perp$ preserves identity structure that an independently trained face-recognition model agrees with, ruling out the circularity concern.}
\label{tab:supp_arcface_idsep}
\small
\begin{tabular}{@{}lrrr@{}}
\toprule
Partition & $n$ & mean cos(arc) & mean cos($\mathbf{z}_\perp$) \\
\midrule
same target identity        & 11{,}445       & $+0.851$ & $+0.232$ \\
different target identity   & 1{,}643{,}845  & $+0.009$ & $+0.053$ \\
\midrule
\textbf{separation $\Delta$ (same$-$different)} & --- & $\mathbf{+0.843}$ & $\mathbf{+0.179}$ \\
\bottomrule
\end{tabular}
\end{table}

\paragraph{Artefacts and reproducibility.}
The pair manifest (\path{ff_pair_manifest.jsonl}, 1820 rows), the two per-pair embedding matrices (\path{zperp_embeddings.npz}, \path{arcface_embeddings.npz}), and the full pair-pair similarity table (\path{pair_similarities.csv}) are released under \texttt{arcface/} in the accompanying supplementary archive, alongside the summary in \path{orthogonality_summary.md}.

\section{Extended Ablation Study}
\label{sec:supp_ablation}

The ablation table in the main paper (Table~\ref{tab:ablation}) collapses the per-component sweep for compactness.
This appendix reports the loss-weight inventory consolidated from main-paper Sec.~\ref{subsec:objective} and the \gtl / \opa component-sensitivity rows (\emph{no hard RR negatives}, \emph{no small-transform penalty}, per-video $\bar{\mathbf{v}}$ vs per-pair $\mathbf{v}$) that were removed from the main Table~\ref{tab:ablation} for space.
All rows use the single training framework of main-paper Table~\ref{tab:cross_results} (macro-best checkpoint; seeds 42, 137, 1729, except the per-video $\bar{\mathbf{v}}$ row at seed 42 only).

\defFloatSpacing{tabSuppLossWeights}
\begin{table}[h!]
\centering
\fsCapAbove{tabSuppLossWeights}
\caption{\textbf{Loss-weight inventory.} All static values used throughout the paper; no per-dataset tuning. Outer scalars $\alpha,\beta,\gamma$ appear in main-paper Eq.~\ref{eq:total}; the inner constants are reused from the section in which each was introduced (\opa: \S\ref{subsec:opa}; \gtl: \S\ref{subsec:gtl}; GSD: \S\ref{subsec:gsd}).}
\label{tab:supp_loss_weights}
\fsCapBelow{tabSuppLossWeights}
\small
\setlength{\tabcolsep}{6pt}
\begin{tabular}{@{}l l l l@{}}
\toprule
Component & Symbol & Value & Introduced \\
\midrule
\opa outer weight              & $\alpha$                 & $1.0$              & \S\ref{subsec:objective}, Eq.~\ref{eq:total} \\
Uniformity outer weight        & $\beta$                  & $0.5$              & \S\ref{subsec:objective}, Eq.~\ref{eq:total} \\
\gtl outer weight              & $\gamma$                 & $1.0$              & \S\ref{subsec:objective}, Eq.~\ref{eq:total} \\
\opa axis-margin               & $m$                      & $1.5$              & \S\ref{subsec:opa}, Eq.~\ref{eq:opa_loss} \\
\gtl small-transform penalty   & $\lambda_{\text{small}}$ & $0.2$              & \S\ref{subsec:gtl}, Eq.~\ref{eq:gtl_loss} \\
\gtl angular/Euclidean mixing  & $c$                      & $0.01$             & \S\ref{subsec:gtl}, Eq.~\ref{eq:gtl_loss} \\
GSD within-video concentration & $\eta_{\text{c}}$        & $5\!\times\!10^{-3}$ & \S\ref{subsec:gsd}, Eq.~\ref{eq:gsd_conc_coh} \\
GSD within-video coherence     & $\eta_{\text{h}}$        & $5\!\times\!10^{-3}$ & \S\ref{subsec:gsd}, Eq.~\ref{eq:gsd_conc_coh} \\
GSD between-video dispersion   & $\eta_{\text{d}}$        & $2\!\times\!10^{-3}$ & \S\ref{subsec:gsd}, Eq.~\ref{eq:gsd_disp} \\
\bottomrule
\end{tabular}
\end{table}
 
\begin{table}[h!]
\centering
\fsCapAbove{tabSuppAblationSensitivity}
\caption{\textbf{GTL / OPA component-sensitivity ablations (Video-AUC \%).} Same training framework as main-paper Table~\ref{tab:ablation}. Rows are grouped by adapter tier, each with its own anchor, because the two tiers are not comparable to one another: the LN-tier anchor is the 3-seed Full method (91.97) and the GSD-LoRA-tier anchor is seed-42 GSD-LoRA (92.48, Table~\ref{tab:supp_full_breakdown_macro}). The sensitivity rows each remove one GTL sub-component or swap the OPA pair-direction. Seeds 42/137/1729, except the two GSD-LoRA-tier rows (s42 only, run on separate hardware). Per-row $n$ annotated.}
\label{tab:supp_ablation_sensitivity}
\fsCapBelow{tabSuppAblationSensitivity}
\small
\setlength{\tabcolsep}{6pt}
\begin{tabular}{@{}lccc@{}}
\toprule
Configuration & DSv2 & CDFv3 & MEAN \\
\midrule
\multicolumn{4}{@{}l}{\textit{LN tier}} \\
\quad + OPA + uniformity + GTL (\textbf{Full method}, anchor) & 81.33 & 83.37 & 91.97 {\tiny ($n{=}3$)} \\
\quad \textit{Full method, no hard RR negatives} & 81.80 & 81.26 & 91.76 {\tiny ($n{=}3$)} \\
\quad \textit{Full method, no small-transform penalty} & 82.03 & 81.25 & 91.69 {\tiny ($n{=}3$)} \\
\midrule
\multicolumn{4}{@{}l}{\textit{GSD-LoRA tier} (seed 42 only)} \\
\quad GSD-LoRA, per-pair $\mathbf{v}$ (anchor) & 83.08 & 83.77 & 92.48 {\tiny ($n{=}1$)} \\
\quad \textit{GSD-LoRA, per-video $\bar{\mathbf{v}}$ instead of per-pair $\mathbf{v}$} & 82.25 & 82.59 & 91.80 {\tiny ($n{=}1$)} \\
\bottomrule
\end{tabular}
\end{table}
 
\paragraph{Component-sensitivity reading.}
On the LN tier, removing either \gtl sub-component from the Full method (91.97 MEAN-17) costs under $0.3$ point: dropping the intra-batch hard \RtoR negatives gives 91.76 ($-0.21$) and dropping the small-transform penalty gives 91.69 ($-0.28$, low seed variance $\sigma{=}0.04$).
Neither \gtl sub-component is individually load-bearing on the LN tier; each contributes incrementally, and the full geometric stack is what realises the headline gain.
On the GSD-LoRA tier, replacing the per-pair OPA direction $\mathbf{v}$ with the per-video mean $\bar{\mathbf{v}}$ costs $0.68$ at fixed seed 42 (92.48 $\to$ 91.80): activating the residual-consistency term that is identically zero under per-pair $\mathbf{v}$ (Sec.~\ref{subsec:opa}) does not help, confirming the per-pair formulation as the better design once the set-level GSD term is active.

\section{Hyperparameter Sensitivity Sweeps}
\label{sec:supp_hyperparam_sweeps}

The \method{} objective introduces four numeric hyperparameters beyond the standard cross-entropy weights: the \opa hinge margin $m$, the \gtl small-transform penalty $\lambda_{\text{small}}$, the \gtl mixing constant $c$ used to evaluate the geodesic transformation, and the GSD concentration/coherence weight $\eta_c{=}\eta_h$ (the dispersion weight is held fixed throughout).
Table~\ref{tab:supp_hyperparam_sweeps} reports a directional single-seed sweep of all four under a truncated 10-epoch training budget (vs.~25 epochs for the headline LN runs in Table~\ref{tab:cross_results} and the same 10-epoch budget for the GSD-LoRA tier); the default value for each hyperparameter is not re-run at the truncated budget and is listed in the panel header rather than as a row.
The single exception is \emph{sweep\_eta\_1e-3}, whose config carried $\texttt{nEpochs}{=}25$ rather than the intended $10$; the run was stopped at epoch~11 to honour the 10-epoch sweep contract, and its macro-best checkpoint sits at epoch~6 -- inside that contract -- so the reported number is comparable to the other sweep rows.
Because the LN-tier sweeps run for fewer epochs than the headline LN configuration, absolute MEAN-17 numbers in panels 1--3 should be read as a \emph{within-panel} comparison and not against Table~\ref{tab:cross_results}; the GSD-LoRA panel is at the same epoch budget as the headline GSD-LoRA run and can be read against the seed-42 GSD-LoRA result.

\paragraph{Hinge margin $m$.}
Across $m\in\{0.5,\,1.0,\,2.0,\,3.0\}$ the cross-domain MEAN-17 stays in the band $90.58$--$91.01$ (range $0.43$).
There is a weak preference for $m\!\approx\!1$--$1.5$ (the peak in the sweep is at $m{=}1.0$ with MEAN-17$=$$91.01$, $0.43$ above the lowest point at $m{=}3.0$ and $0.30$ above $m{=}2.0$), consistent with the default choice of $m{=}1.5$ used in the main runs.
The hinge geometry is therefore insensitive to the precise margin within the $[0.5,\,3.0]$ range.

\paragraph{Small-transform penalty $\lambda_{\text{small}}$.}
Sweeping $\lambda_{\text{small}}\in\{0.05,\,0.5\}$ moves MEAN-17 between $90.76$ and $91.07$ (range $0.31$).
Halving the weight from the default $0.2$ to $0.05$ gives the highest FF-FSH of the sweep ($88.98$), while increasing it $2.5\times$ to $0.5$ trades FF-FSH ($-1.71$) for MEAN-17 ($+0.31$), indicating that the small-transform term primarily controls a stability/transfer trade-off rather than a level shift.
The full-removal ablation agrees: dropping the term entirely costs only $0.28$ MEAN-17 ($91.97\to91.69$ at $\sigma{=}0.04$, $n{=}3$; Table~\ref{tab:supp_ablation_sensitivity}), a small and stable effect.

\paragraph{GTL mixing constant $c$.}
The mixing constant $c$ was held fixed at $0.01$ throughout the main runs.
Sweeping it across two orders of magnitude --- $c\!\in\!\{0.001,\,0.1\}$ --- leaves MEAN-17 in a $0.19$-point band ($91.02$--$91.21$), with the higher value (and a fast-converging $c{=}0.1$ run that peaks at epoch $7$ rather than $9$) slightly preferred.
The geodesic mixing constant is therefore not a load-bearing hyperparameter; the choice $c{=}0.01$ in the main runs is robust.

\paragraph{GSD concentration / coherence weight $\eta_c{=}\eta_h$.}
The GSD-tier sweep is at the same 10-epoch budget as the headline GSD-LoRA run and so can be compared directly: seed-42 GSD-LoRA at the default $\eta_c{=}\eta_h{=}5\!\times\!10^{-3}$ scores MEAN-17$=$$92.48$ (epoch $7$, Table~\ref{tab:supp_full_breakdown_macro}).
Increasing the weight $4\times$ to $2\!\times\!10^{-2}$ accelerates convergence (peak at epoch $3$) and lifts MEAN-17 to $93.06$ and FF-FSH to $90.72$, a $+0.58$ MEAN-17 gain over the default at the same seed and budget.
Decreasing the weight $5\times$ to $1\!\times\!10^{-3}$ slows training (peak at epoch $6$) and lands at MEAN-17$=$$92.33$, $0.15$ below the default.
The single-seed signal therefore suggests the GSD-LoRA result is not yet weight-saturated and that the released configuration is mildly conservative; a multi-seed re-sweep at $\eta_c{=}\eta_h{=}2\!\times\!10^{-2}$ is the most promising follow-up suggested by these data.

\begin{table}[t]
\centering
\fsCapAbove{tabSuppHyperparamSweeps}
\caption{\textbf{Single-seed sensitivity sweeps for the four numeric loss
hyperparameters introduced by \method{}.} Each row is a single sweep run
at seed 42 under a truncated 10-epoch budget (vs.~25 epochs in the
headline LN runs and the same 10-epoch budget for the GSD-LoRA tier);
absolute MEAN-17 values are therefore directional and should be
compared \emph{within} a panel rather than against Tables~\ref{tab:cross_results}
or~\ref{tab:ablation}. Default values are listed in the panel header but
were not re-run at the 10-epoch budget (the seed-42 default is the
corresponding headline run). The hinge margin $m$, small-transform
penalty $\lambda_{\text{small}}$, and GTL mixing constant $c$ are swept
on the LN tier; the GSD concentration / coherence weight
$\eta_c{=}\eta_h$ (the GSD dispersion weight is held fixed at
$2\!\times\!10^{-3}$) is swept on the GSD-LoRA tier. Each panel reports
the epoch at which the macro-best checkpoint was selected, Video-AUC on
the held-out FF-FSH-C23 head, and the cross-domain MEAN-17 over the 17
external test sets. The five in-domain FF++-c23 test sets (FF-ORIGINAL,
FF-DF, FF-F2F, FF-FS, FF-NT) sit in the $98.3$--$99.4$ Video-AUC range
across all 10 sweep runs and are omitted from the table to save space.}
\label{tab:supp_hyperparam_sweeps}
\fsCapBelow{tabSuppHyperparamSweeps}
\small
\setlength{\tabcolsep}{6pt}
\begin{tabular}{@{}lccc@{}}
\toprule
Value & Epoch & FF-FSH & MEAN-17 \\
\midrule
\multicolumn{4}{@{}l}{\textbf{Hinge margin $m$} \quad (tier: LN; default $= 1.5$)} \\
\midrule
\quad 0.5 & ep 09 & 88.55 & 90.72 \\
\quad 1.0 & ep 09 & 88.35 & 91.01 \\
\quad 2.0 & ep 09 & 87.65 & 90.71 \\
\quad 3.0 & ep 09 & 87.80 & 90.58 \\
\midrule
\multicolumn{4}{@{}l}{\textbf{Small-transform penalty $\lambda_{\text{small}}$} \quad (tier: LN; default $= 0.2$)} \\
\midrule
\quad 0.05 & ep 07 & 88.98 & 90.76 \\
\quad 0.5 & ep 07 & 87.27 & 91.07 \\
\midrule
\multicolumn{4}{@{}l}{\textbf{GTL mixing constant $c$} \quad (tier: LN; default $= 0.01$)} \\
\midrule
\quad 0.001 & ep 09 & 88.24 & 91.02 \\
\quad 0.1 & ep 07 & 88.36 & 91.21 \\
\midrule
\multicolumn{4}{@{}l}{\textbf{GSD weight $\eta_c{=}\eta_h$} \quad (tier: GSD-LoRA; default $= 5\times10^{-3}$)} \\
\midrule
\quad $1\times10^{-3}$ & ep 06 & 89.72 & 92.33 \\
\quad $2\times10^{-2}$ & ep 03 & 90.72 & 93.06 \\
\bottomrule
\end{tabular}
\end{table}

\section{Training Dynamics: Per-seed Peak and Plateau Summary}
\label{sec:supp_training_dynamics}

\begin{figure}[!htbp]
    \centering
    \includegraphics[width=0.9\linewidth]{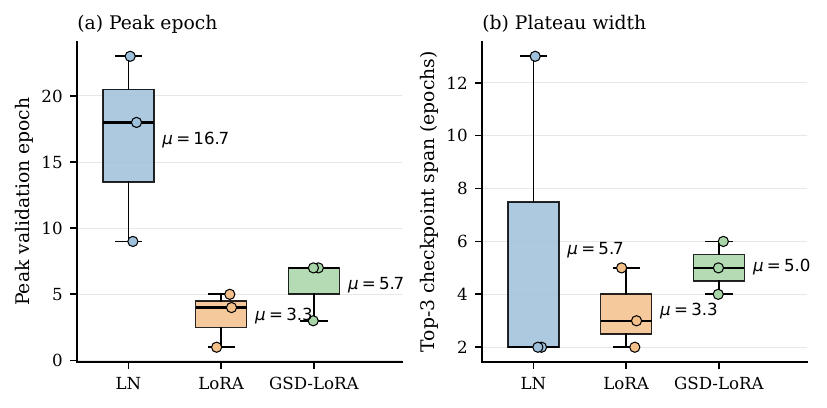}
    \fsCapAbove{figSuppBoxSummary}
    \caption{\textbf{Per-seed peak-epoch and plateau-width summary.} \textbf{(a)}~Peak validation epoch per variant (boxplot $+$ per-seed scatter). \textbf{(b)}~Top-3 checkpoint span (epochs between the highest and lowest of each run's three best validation epochs), used as a proxy for the high-validation-AUC plateau width. Three seeds per variant; $\mu$ annotations show the mean.}
    \label{fig:supp_box_summary}
    \fsCapBelow{figSuppBoxSummary}
\end{figure}

The main paper's Fig.~\ref{fig:val_curves} shows the macro validation Video-AUC trajectory for each variant.
Figure~\ref{fig:supp_box_summary} complements this with the per-seed peak-epoch distribution and the top-3 checkpoint span (used as a proxy for the high-validation-AUC plateau width).
LN tuning peaks late (median epoch 18, range $9$--$23$) but holds its peak for a long window.
Plain LoRA peaks early (mean epoch $3.3$) with a narrow top-3 plateau ($3.3$ epochs); GSD-LoRA delays the peak to epoch $5.7$ and widens the plateau to $5.0$ epochs, a $1.5\times$ improvement at the same parameter budget.

\fsBefore{figSuppBoxSummary}
\fsAfter{figSuppBoxSummary}

\section{Full Per-Run Checkpoint Breakdown}
\label{sec:supp_full_breakdown}

This appendix exposes the raw per-seed per-dataset Video-AUC table behind every cell of Tables~\ref{tab:cross_results} and~\ref{tab:supp_checkpoint_policy}, for transparency and to let reviewers verify the cross-domain mean numbers cell-for-cell.
Each of the 9 main runs (3 configurations $\times$ 3 seeds) was tested at four checkpoints, one per selection criterion: \emph{macro-best} (default, used in main-paper Table~\ref{tab:cross_results}) and three single-val criteria (DSv2-VAL only, CDFv3-VAL only, FF-FSH-VAL only).
Tables~\ref{tab:supp_full_breakdown_macro}--\ref{tab:supp_full_breakdown_ffsh} report all 36 (run, criterion) combinations.
Three-seed means with $\pm\sigma$ are shown in bold beneath each method group.
Run labels carry the \emph{checkpoint} epoch, which is the epoch of the selected \texttt{macro-best} checkpoint file; this coincides with the validation peak for eight of the nine runs, the exception being LN (s137), whose checkpoint is written at epoch 24 while its validation peak sits at epoch 23 ($85.56$ vs.\ $85.55$). Main-paper Table~\ref{tab:training_stability} reports validation-curve peaks ($\{18,23,9\}$ for LN), so the two are consistent by construction.
The Unseen column is the cross-17 mean for that run and ties to Table~\ref{tab:supp_checkpoint_policy} exactly.

\paragraph{Reading guide.}
The macro-best block reproduces main-paper Table~\ref{tab:cross_results} cell-for-cell.
The three single-val blocks let the reader probe how each cross-domain win depends on the validation set used for early stopping: under FF-FSH-VAL-only selection in particular, DSv2 and CDFv3 are genuinely zero-shot (the model has never seen any DSv2- or CDFv3-family data in either training or checkpoint selection), and the cross-domain mean stays at 92.74\,$\pm$\,0.32 for GSD-LoRA --- $1.16$ above GenD$^{\mathrm{DINO}}$ on the same pipeline.
Every cell in these tables is computed directly from the \path{test_results.json} files shipped at \path{logs/main/<run>/test/<epochNN_best_{macro,dsv2,cdfv3,ffsh}>/test_results.json}; the headline tables (main paper Tables~\ref{tab:indomain_results}, \ref{tab:cross_results} and supp Table~\ref{tab:supp_checkpoint_policy}) can be reprinted from the shipped archive by running \texttt{python3 reproduce\_tables.py} at the archive root.

\begin{landscape}
\begin{table*}[p]
\centering
\caption{\textbf{Per-seed per-dataset Video-AUC (\%): selection criterion \emph{macro-best (default)}.} All 9 main runs (3 configurations $\times$ 3 seeds) at the checkpoint chosen under the named selection rule. The final column (\emph{Unseen}) is the across-17-dataset mean. Bold rows are 3-seed means $\pm\sigma$.}
\label{tab:supp_full_breakdown_macro}
\scriptsize
\setlength{\tabcolsep}{2.4pt}
\resizebox{\linewidth}{!}{\begin{tabular}{@{}lcccccccccccccccccc@{}}
\toprule
Run & FFSH & CDFv1 & CDFv2 & CDFv3 & DeFo & DFD & DSv1 & DSv2 & DFDCP & DFDC & DFDM & FAVC & FFIW & IDFv1 & KoDF & PGF & UADFV & Unseen \\
\midrule
LN (s42, epoch18) & 88.86 & 94.23 & 93.96 & 83.49 & 97.23 & 98.24 & 86.30 & 81.15 & 87.28 & 86.30 & 99.88 & 97.97 & 90.52 & 97.12 & 92.13 & 91.24 & 97.83 & 91.98 \\
LN (s137, epoch24) & 88.92 & 94.23 & 94.27 & 84.28 & 97.39 & 98.09 & 86.30 & 81.63 & 87.60 & 85.89 & 99.89 & 97.88 & 90.27 & 96.99 & 91.96 & 91.45 & 97.96 & 92.06 \\
LN (s1729, epoch09) & 88.83 & 94.52 & 93.81 & 82.32 & 97.72 & 98.12 & 85.83 & 81.20 & 87.55 & 86.39 & 99.83 & 97.95 & 90.40 & 96.99 & 91.37 & 90.39 & 98.29 & 91.85 \\
\textbf{LN mean ($n{=}3$)} & 88.87\,{\tiny$\pm$0.04} & 94.33\,{\tiny$\pm$0.17} & 94.01\,{\tiny$\pm$0.24} & 83.37\,{\tiny$\pm$0.99} & 97.45\,{\tiny$\pm$0.25} & 98.15\,{\tiny$\pm$0.08} & 86.14\,{\tiny$\pm$0.27} & 81.33\,{\tiny$\pm$0.26} & 87.47\,{\tiny$\pm$0.17} & 86.19\,{\tiny$\pm$0.27} & 99.87\,{\tiny$\pm$0.03} & 97.93\,{\tiny$\pm$0.05} & 90.39\,{\tiny$\pm$0.12} & 97.04\,{\tiny$\pm$0.08} & 91.82\,{\tiny$\pm$0.40} & 91.03\,{\tiny$\pm$0.56} & 98.03\,{\tiny$\pm$0.24} & 91.97\,{\tiny$\pm$0.10} \\
\midrule
LoRA (s42, epoch04) & 92.58 & 94.27 & 93.57 & 84.02 & 99.83 & 97.05 & 88.29 & 82.75 & 91.68 & 87.02 & 99.87 & 98.30 & 87.88 & 97.06 & 89.07 & 95.24 & 97.75 & 92.72 \\
LoRA (s137, epoch05) & 90.40 & 92.49 & 92.62 & 85.31 & 99.32 & 96.65 & 89.48 & 83.07 & 91.16 & 84.98 & 99.61 & 97.79 & 87.44 & 97.09 & 89.17 & 94.50 & 97.04 & 92.24 \\
LoRA (s1729, epoch01) & 91.85 & 96.43 & 95.11 & 84.01 & 99.04 & 98.44 & 87.94 & 81.29 & 90.42 & 86.82 & 99.97 & 99.12 & 88.98 & 97.83 & 89.47 & 94.59 & 97.79 & 92.89 \\
\textbf{LoRA mean ($n{=}3$)} & 91.61\,{\tiny$\pm$1.11} & 94.40\,{\tiny$\pm$1.98} & 93.77\,{\tiny$\pm$1.26} & 84.45\,{\tiny$\pm$0.75} & 99.40\,{\tiny$\pm$0.40} & 97.38\,{\tiny$\pm$0.94} & 88.57\,{\tiny$\pm$0.81} & 82.37\,{\tiny$\pm$0.95} & 91.09\,{\tiny$\pm$0.63} & 86.27\,{\tiny$\pm$1.12} & 99.81\,{\tiny$\pm$0.19} & 98.40\,{\tiny$\pm$0.67} & 88.10\,{\tiny$\pm$0.80} & 97.33\,{\tiny$\pm$0.43} & 89.24\,{\tiny$\pm$0.21} & 94.78\,{\tiny$\pm$0.40} & 97.53\,{\tiny$\pm$0.42} & 92.62\,{\tiny$\pm$0.33} \\
\midrule
GSD-LoRA (s42, epoch07) & 92.05 & 92.28 & 92.22 & 83.77 & 99.44 & 97.89 & 89.80 & 83.08 & 90.58 & 85.82 & 99.79 & 98.50 & 88.07 & 97.53 & 88.79 & 94.68 & 97.96 & 92.48 \\
GSD-LoRA (s137, epoch03) & 92.02 & 94.61 & 93.79 & 84.75 & 98.87 & 97.82 & 91.31 & 86.23 & 91.31 & 86.02 & 99.92 & 99.35 & 89.79 & 98.43 & 89.77 & 93.74 & 96.42 & 93.18 \\
GSD-LoRA (s1729, epoch07) & 93.13 & 93.93 & 94.01 & 86.30 & 99.42 & 97.24 & 88.62 & 83.81 & 92.41 & 87.19 & 99.76 & 97.64 & 88.53 & 96.87 & 88.67 & 93.32 & 96.58 & 92.79 \\
\textbf{GSD-LoRA mean ($n{=}3$)} & 92.40\,{\tiny$\pm$0.63} & 93.60\,{\tiny$\pm$1.20} & 93.34\,{\tiny$\pm$0.98} & 84.94\,{\tiny$\pm$1.28} & 99.24\,{\tiny$\pm$0.32} & 97.65\,{\tiny$\pm$0.36} & 89.91\,{\tiny$\pm$1.35} & 84.37\,{\tiny$\pm$1.65} & 91.43\,{\tiny$\pm$0.92} & 86.34\,{\tiny$\pm$0.74} & 99.82\,{\tiny$\pm$0.08} & 98.50\,{\tiny$\pm$0.85} & 88.79\,{\tiny$\pm$0.89} & 97.61\,{\tiny$\pm$0.78} & 89.07\,{\tiny$\pm$0.60} & 93.92\,{\tiny$\pm$0.70} & 96.99\,{\tiny$\pm$0.85} & 92.82\,{\tiny$\pm$0.35} \\
\bottomrule
\end{tabular}}
\end{table*}
\end{landscape}

\begin{landscape}
\begin{table*}[p]
\centering
\caption{\textbf{Per-seed per-dataset Video-AUC (\%): selection criterion \emph{DSv2-VAL only}.} All 9 main runs (3 configurations $\times$ 3 seeds) at the checkpoint chosen under the named selection rule. The final column (\emph{Unseen}) is the across-17-dataset mean. Bold rows are 3-seed means $\pm\sigma$.}
\label{tab:supp_full_breakdown_dsv2}
\scriptsize
\setlength{\tabcolsep}{2.4pt}
\resizebox{\linewidth}{!}{\begin{tabular}{@{}lcccccccccccccccccc@{}}
\toprule
Run & FFSH & CDFv1 & CDFv2 & CDFv3 & DeFo & DFD & DSv1 & DSv2 & DFDCP & DFDC & DFDM & FAVC & FFIW & IDFv1 & KoDF & PGF & UADFV & Unseen \\
\midrule
LN (s42, epoch04) & 88.07 & 92.87 & 92.31 & 77.63 & 96.83 & 97.00 & 83.77 & 80.23 & 88.50 & 85.79 & 99.63 & 97.33 & 88.41 & 95.38 & 89.75 & 85.69 & 98.67 & 90.46 \\
LN (s137, epoch05) & 87.68 & 93.80 & 92.45 & 78.70 & 97.25 & 97.59 & 83.77 & 80.06 & 88.69 & 86.11 & 99.77 & 97.52 & 89.05 & 95.70 & 90.02 & 87.22 & 98.67 & 90.83 \\
LN (s1729, epoch03) & 88.62 & 92.49 & 91.64 & 75.64 & 97.52 & 96.42 & 82.42 & 80.28 & 88.92 & 85.61 & 99.41 & 96.69 & 87.96 & 94.58 & 88.75 & 83.71 & 98.83 & 89.97 \\
\textbf{LN mean ($n{=}3$)} & 88.12\,{\tiny$\pm$0.47} & 93.05\,{\tiny$\pm$0.68} & 92.13\,{\tiny$\pm$0.43} & 77.32\,{\tiny$\pm$1.55} & 97.20\,{\tiny$\pm$0.34} & 97.00\,{\tiny$\pm$0.59} & 83.32\,{\tiny$\pm$0.78} & 80.19\,{\tiny$\pm$0.11} & 88.70\,{\tiny$\pm$0.21} & 85.84\,{\tiny$\pm$0.25} & 99.61\,{\tiny$\pm$0.18} & 97.18\,{\tiny$\pm$0.43} & 88.47\,{\tiny$\pm$0.55} & 95.22\,{\tiny$\pm$0.58} & 89.50\,{\tiny$\pm$0.67} & 85.54\,{\tiny$\pm$1.76} & 98.72\,{\tiny$\pm$0.10} & 90.42\,{\tiny$\pm$0.43} \\
\midrule
LoRA (s42, epoch04) & 92.58 & 94.27 & 93.57 & 84.02 & 99.83 & 97.05 & 88.29 & 82.75 & 91.68 & 87.02 & 99.87 & 98.30 & 87.88 & 97.06 & 89.07 & 95.24 & 97.75 & 92.72 \\
LoRA (s137, epoch05) & 90.40 & 92.49 & 92.62 & 85.31 & 99.32 & 96.65 & 89.48 & 83.07 & 91.16 & 84.98 & 99.61 & 97.79 & 87.44 & 97.09 & 89.17 & 94.50 & 97.04 & 92.24 \\
LoRA (s1729, epoch02) & 91.40 & 94.19 & 92.68 & 84.42 & 99.14 & 97.91 & 89.67 & 82.98 & 91.01 & 86.34 & 99.80 & 98.31 & 87.82 & 96.60 & 88.76 & 93.49 & 97.79 & 92.49 \\
\textbf{LoRA mean ($n{=}3$)} & 91.46\,{\tiny$\pm$1.09} & 93.65\,{\tiny$\pm$1.01} & 92.96\,{\tiny$\pm$0.54} & 84.58\,{\tiny$\pm$0.66} & 99.43\,{\tiny$\pm$0.36} & 97.20\,{\tiny$\pm$0.64} & 89.15\,{\tiny$\pm$0.75} & 82.94\,{\tiny$\pm$0.16} & 91.28\,{\tiny$\pm$0.35} & 86.11\,{\tiny$\pm$1.04} & 99.76\,{\tiny$\pm$0.13} & 98.13\,{\tiny$\pm$0.29} & 87.71\,{\tiny$\pm$0.24} & 96.92\,{\tiny$\pm$0.28} & 89.00\,{\tiny$\pm$0.22} & 94.41\,{\tiny$\pm$0.88} & 97.53\,{\tiny$\pm$0.42} & 92.48\,{\tiny$\pm$0.24} \\
\midrule
GSD-LoRA (s42, epoch02) & 90.27 & 96.05 & 93.97 & 84.07 & 98.78 & 98.71 & 88.84 & 83.55 & 90.78 & 87.65 & 99.92 & 98.63 & 89.26 & 97.16 & 91.19 & 94.28 & 98.50 & 93.04 \\
GSD-LoRA (s137, epoch03) & 92.02 & 94.61 & 93.79 & 84.75 & 98.87 & 97.82 & 91.31 & 86.23 & 91.31 & 86.02 & 99.92 & 99.35 & 89.79 & 98.43 & 89.77 & 93.74 & 96.42 & 93.18 \\
GSD-LoRA (s1729, epoch11) & 91.95 & 92.02 & 90.23 & 84.23 & 99.52 & 96.74 & 89.93 & 83.34 & 90.88 & 85.59 & 99.56 & 98.01 & 87.65 & 97.07 & 87.99 & 94.57 & 97.88 & 92.19 \\
\textbf{GSD-LoRA mean ($n{=}3$)} & 91.41\,{\tiny$\pm$0.99} & 94.23\,{\tiny$\pm$2.04} & 92.66\,{\tiny$\pm$2.11} & 84.35\,{\tiny$\pm$0.35} & 99.06\,{\tiny$\pm$0.40} & 97.76\,{\tiny$\pm$0.99} & 90.03\,{\tiny$\pm$1.24} & 84.37\,{\tiny$\pm$1.61} & 90.99\,{\tiny$\pm$0.28} & 86.42\,{\tiny$\pm$1.08} & 99.80\,{\tiny$\pm$0.21} & 98.66\,{\tiny$\pm$0.67} & 88.90\,{\tiny$\pm$1.11} & 97.55\,{\tiny$\pm$0.76} & 89.65\,{\tiny$\pm$1.60} & 94.19\,{\tiny$\pm$0.42} & 97.60\,{\tiny$\pm$1.07} & 92.80\,{\tiny$\pm$0.54} \\
\bottomrule
\end{tabular}}
\end{table*}
\end{landscape}

\begin{landscape}
\begin{table*}[p]
\centering
\caption{\textbf{Per-seed per-dataset Video-AUC (\%): selection criterion \emph{CDFv3-VAL only}.} All 9 main runs (3 configurations $\times$ 3 seeds) at the checkpoint chosen under the named selection rule. The final column (\emph{Unseen}) is the across-17-dataset mean. Bold rows are 3-seed means $\pm\sigma$.}
\label{tab:supp_full_breakdown_cdfv3}
\scriptsize
\setlength{\tabcolsep}{2.4pt}
\resizebox{\linewidth}{!}{\begin{tabular}{@{}lcccccccccccccccccc@{}}
\toprule
Run & FFSH & CDFv1 & CDFv2 & CDFv3 & DeFo & DFD & DSv1 & DSv2 & DFDCP & DFDC & DFDM & FAVC & FFIW & IDFv1 & KoDF & PGF & UADFV & Unseen \\
\midrule
LN (s42, epoch15) & 88.66 & 94.10 & 93.84 & 83.44 & 97.15 & 98.23 & 86.31 & 81.02 & 87.31 & 86.18 & 99.88 & 97.95 & 90.41 & 97.12 & 92.02 & 91.03 & 97.88 & 91.91 \\
LN (s137, epoch23) & 88.92 & 94.23 & 94.26 & 84.26 & 97.40 & 98.09 & 86.29 & 81.63 & 87.64 & 85.90 & 99.89 & 97.88 & 90.26 & 96.98 & 91.95 & 91.44 & 97.96 & 92.06 \\
LN (s1729, epoch23) & 88.80 & 94.74 & 94.25 & 83.43 & 97.69 & 98.17 & 86.11 & 80.97 & 87.05 & 85.74 & 99.89 & 97.98 & 90.34 & 97.24 & 91.92 & 91.55 & 98.21 & 92.01 \\
\textbf{LN mean ($n{=}3$)} & 88.79\,{\tiny$\pm$0.13} & 94.35\,{\tiny$\pm$0.34} & 94.12\,{\tiny$\pm$0.24} & 83.71\,{\tiny$\pm$0.48} & 97.42\,{\tiny$\pm$0.27} & 98.16\,{\tiny$\pm$0.07} & 86.24\,{\tiny$\pm$0.11} & 81.20\,{\tiny$\pm$0.37} & 87.33\,{\tiny$\pm$0.30} & 85.94\,{\tiny$\pm$0.22} & 99.89\,{\tiny$\pm$0.01} & 97.94\,{\tiny$\pm$0.05} & 90.34\,{\tiny$\pm$0.07} & 97.11\,{\tiny$\pm$0.13} & 91.97\,{\tiny$\pm$0.05} & 91.34\,{\tiny$\pm$0.28} & 98.01\,{\tiny$\pm$0.17} & 91.99\,{\tiny$\pm$0.07} \\
\midrule
LoRA (s42, epoch04) & 92.58 & 94.27 & 93.57 & 84.02 & 99.83 & 97.05 & 88.29 & 82.75 & 91.68 & 87.02 & 99.87 & 98.30 & 87.88 & 97.06 & 89.07 & 95.24 & 97.75 & 92.72 \\
LoRA (s137, epoch04) & 88.38 & 92.02 & 92.07 & 83.88 & 99.12 & 96.80 & 89.43 & 81.88 & 91.69 & 85.79 & 99.76 & 98.35 & 85.20 & 97.13 & 88.43 & 94.62 & 96.83 & 91.85 \\
LoRA (s1729, epoch02) & 91.40 & 94.19 & 92.68 & 84.42 & 99.14 & 97.91 & 89.67 & 82.98 & 91.01 & 86.34 & 99.80 & 98.31 & 87.82 & 96.60 & 88.76 & 93.49 & 97.79 & 92.49 \\
\textbf{LoRA mean ($n{=}3$)} & 90.79\,{\tiny$\pm$2.17} & 93.49\,{\tiny$\pm$1.27} & 92.77\,{\tiny$\pm$0.76} & 84.11\,{\tiny$\pm$0.28} & 99.36\,{\tiny$\pm$0.40} & 97.25\,{\tiny$\pm$0.58} & 89.13\,{\tiny$\pm$0.74} & 82.54\,{\tiny$\pm$0.58} & 91.46\,{\tiny$\pm$0.39} & 86.38\,{\tiny$\pm$0.62} & 99.81\,{\tiny$\pm$0.05} & 98.32\,{\tiny$\pm$0.03} & 86.97\,{\tiny$\pm$1.53} & 96.93\,{\tiny$\pm$0.29} & 88.75\,{\tiny$\pm$0.32} & 94.45\,{\tiny$\pm$0.88} & 97.46\,{\tiny$\pm$0.54} & 92.35\,{\tiny$\pm$0.45} \\
\midrule
GSD-LoRA (s42, epoch15) & 89.63 & 94.61 & 93.04 & 86.21 & 99.60 & 96.55 & 88.26 & 82.14 & 91.84 & 85.77 & 99.69 & 97.76 & 86.40 & 95.40 & 87.14 & 94.04 & 97.38 & 92.09 \\
GSD-LoRA (s137, epoch03) & 92.02 & 94.61 & 93.79 & 84.75 & 98.87 & 97.82 & 91.31 & 86.23 & 91.31 & 86.02 & 99.92 & 99.35 & 89.79 & 98.43 & 89.77 & 93.74 & 96.42 & 93.18 \\
GSD-LoRA (s1729, epoch07) & 93.13 & 93.93 & 94.01 & 86.30 & 99.42 & 97.24 & 88.62 & 83.81 & 92.41 & 87.19 & 99.76 & 97.64 & 88.53 & 96.87 & 88.67 & 93.32 & 96.58 & 92.79 \\
\textbf{GSD-LoRA mean ($n{=}3$)} & 91.59\,{\tiny$\pm$1.79} & 94.38\,{\tiny$\pm$0.39} & 93.61\,{\tiny$\pm$0.51} & 85.75\,{\tiny$\pm$0.87} & 99.30\,{\tiny$\pm$0.38} & 97.20\,{\tiny$\pm$0.63} & 89.40\,{\tiny$\pm$1.67} & 84.06\,{\tiny$\pm$2.06} & 91.85\,{\tiny$\pm$0.55} & 86.33\,{\tiny$\pm$0.75} & 99.79\,{\tiny$\pm$0.12} & 98.25\,{\tiny$\pm$0.95} & 88.24\,{\tiny$\pm$1.71} & 96.90\,{\tiny$\pm$1.52} & 88.52\,{\tiny$\pm$1.32} & 93.70\,{\tiny$\pm$0.36} & 96.79\,{\tiny$\pm$0.51} & 92.69\,{\tiny$\pm$0.56} \\
\bottomrule
\end{tabular}}
\end{table*}
\end{landscape}

\begin{landscape}
\begin{table*}[p]
\centering
\caption{\textbf{Per-seed per-dataset Video-AUC (\%): selection criterion \emph{FF-FSH-VAL only}.} All 9 main runs (3 configurations $\times$ 3 seeds) at the checkpoint chosen under the named selection rule. The final column (\emph{Unseen}) is the across-17-dataset mean. Bold rows are 3-seed means $\pm\sigma$.}
\label{tab:supp_full_breakdown_ffsh}
\scriptsize
\setlength{\tabcolsep}{2.4pt}
\resizebox{\linewidth}{!}{\begin{tabular}{@{}lcccccccccccccccccc@{}}
\toprule
Run & FFSH & CDFv1 & CDFv2 & CDFv3 & DeFo & DFD & DSv1 & DSv2 & DFDCP & DFDC & DFDM & FAVC & FFIW & IDFv1 & KoDF & PGF & UADFV & Unseen \\
\midrule
LN (s42, epoch19) & 88.92 & 94.27 & 93.97 & 83.55 & 97.23 & 98.16 & 86.28 & 80.93 & 87.35 & 86.02 & 99.89 & 97.98 & 90.38 & 97.10 & 92.08 & 91.33 & 98.04 & 91.97 \\
LN (s137, epoch22) & 88.93 & 94.27 & 94.26 & 84.24 & 97.42 & 98.08 & 86.26 & 81.57 & 87.65 & 85.92 & 99.89 & 97.87 & 90.27 & 96.97 & 91.94 & 91.41 & 98.04 & 92.06 \\
LN (s1729, epoch13) & 88.79 & 94.65 & 94.11 & 82.85 & 97.85 & 98.19 & 86.20 & 81.27 & 87.16 & 86.08 & 99.84 & 98.03 & 90.54 & 97.22 & 91.75 & 91.41 & 98.25 & 92.01 \\
\textbf{LN mean ($n{=}3$)} & 88.88\,{\tiny$\pm$0.08} & 94.40\,{\tiny$\pm$0.22} & 94.12\,{\tiny$\pm$0.15} & 83.55\,{\tiny$\pm$0.70} & 97.50\,{\tiny$\pm$0.32} & 98.14\,{\tiny$\pm$0.05} & 86.25\,{\tiny$\pm$0.04} & 81.26\,{\tiny$\pm$0.32} & 87.39\,{\tiny$\pm$0.25} & 86.01\,{\tiny$\pm$0.08} & 99.87\,{\tiny$\pm$0.03} & 97.96\,{\tiny$\pm$0.08} & 90.40\,{\tiny$\pm$0.14} & 97.10\,{\tiny$\pm$0.13} & 91.92\,{\tiny$\pm$0.17} & 91.38\,{\tiny$\pm$0.05} & 98.11\,{\tiny$\pm$0.12} & 92.01\,{\tiny$\pm$0.05} \\
\midrule
LoRA (s42, epoch04) & 92.58 & 94.27 & 93.57 & 84.02 & 99.83 & 97.05 & 88.29 & 82.75 & 91.68 & 87.02 & 99.87 & 98.30 & 87.88 & 97.06 & 89.07 & 95.24 & 97.75 & 92.72 \\
LoRA (s137, epoch07) & 89.17 & 89.69 & 89.46 & 82.98 & 99.61 & 95.33 & 90.00 & 82.25 & 89.63 & 84.64 & 99.75 & 98.04 & 85.19 & 96.25 & 87.63 & 94.93 & 97.67 & 91.31 \\
LoRA (s1729, epoch01) & 91.85 & 96.43 & 95.11 & 84.01 & 99.04 & 98.44 & 87.94 & 81.29 & 90.42 & 86.82 & 99.97 & 99.12 & 88.98 & 97.83 & 89.47 & 94.59 & 97.79 & 92.89 \\
\textbf{LoRA mean ($n{=}3$)} & 91.20\,{\tiny$\pm$1.79} & 93.46\,{\tiny$\pm$3.45} & 92.71\,{\tiny$\pm$2.92} & 83.67\,{\tiny$\pm$0.60} & 99.50\,{\tiny$\pm$0.41} & 96.94\,{\tiny$\pm$1.56} & 88.74\,{\tiny$\pm$1.10} & 82.10\,{\tiny$\pm$0.74} & 90.58\,{\tiny$\pm$1.03} & 86.16\,{\tiny$\pm$1.32} & 99.86\,{\tiny$\pm$0.11} & 98.48\,{\tiny$\pm$0.56} & 87.35\,{\tiny$\pm$1.95} & 97.05\,{\tiny$\pm$0.79} & 88.73\,{\tiny$\pm$0.97} & 94.92\,{\tiny$\pm$0.33} & 97.74\,{\tiny$\pm$0.06} & 92.30\,{\tiny$\pm$0.87} \\
\midrule
GSD-LoRA (s42, epoch05) & 93.23 & 94.69 & 94.15 & 84.55 & 99.41 & 97.89 & 89.17 & 83.28 & 92.84 & 86.52 & 99.95 & 98.39 & 86.94 & 97.41 & 90.01 & 94.61 & 98.42 & 93.03 \\
GSD-LoRA (s137, epoch07) & 91.50 & 92.66 & 92.93 & 82.06 & 99.77 & 97.31 & 88.73 & 82.79 & 91.68 & 87.19 & 99.88 & 98.41 & 87.00 & 97.47 & 88.27 & 95.23 & 97.83 & 92.39 \\
GSD-LoRA (s1729, epoch07) & 93.13 & 93.93 & 94.01 & 86.30 & 99.42 & 97.24 & 88.62 & 83.81 & 92.41 & 87.19 & 99.76 & 97.64 & 88.53 & 96.87 & 88.67 & 93.32 & 96.58 & 92.79 \\
\textbf{GSD-LoRA mean ($n{=}3$)} & 92.62\,{\tiny$\pm$0.97} & 93.76\,{\tiny$\pm$1.03} & 93.70\,{\tiny$\pm$0.67} & 84.30\,{\tiny$\pm$2.13} & 99.53\,{\tiny$\pm$0.20} & 97.48\,{\tiny$\pm$0.35} & 88.84\,{\tiny$\pm$0.29} & 83.29\,{\tiny$\pm$0.51} & 92.31\,{\tiny$\pm$0.58} & 86.96\,{\tiny$\pm$0.39} & 99.86\,{\tiny$\pm$0.09} & 98.15\,{\tiny$\pm$0.44} & 87.49\,{\tiny$\pm$0.90} & 97.25\,{\tiny$\pm$0.33} & 88.98\,{\tiny$\pm$0.91} & 94.39\,{\tiny$\pm$0.97} & 97.61\,{\tiny$\pm$0.94} & 92.74\,{\tiny$\pm$0.32} \\
\bottomrule
\end{tabular}}
\end{table*}
\end{landscape}
  

\begin{thebibliography}{27}
\providecommand{\natexlab}[1]{#1}
\providecommand{\url}[1]{\texttt{#1}}
\expandafter\ifx\csname urlstyle\endcsname\relax
  \providecommand{\doi}[1]{doi: #1}\else
  \providecommand{\doi}{doi: \begingroup \urlstyle{rm}\Url}\fi

\bibitem[Barrington et~al.(2026)Barrington, Bohacek, and
  Farid]{barrington2026deepspeak}
Sarah Barrington, Maty Bohacek, and Hany Farid.
\newblock The deepspeak dataset.
\newblock In \emph{Proceedings of the IEEE/CVF Conference on Computer Vision
  and Pattern Recognition (CVPR) Findings}, pages 1893--1902, June 2026.
\newblock URL
  \url{https://openaccess.thecvf.com/content/CVPR2026F/html/Barrington_The_DeepSpeak_Dataset_CVPRF_2026_paper.html}.

\bibitem[Cui et~al.(2025)Cui, Li, Luo, Zhou, and Dong]{cui2025forensics}
Xinjie Cui, Yuezun Li, Ao~Luo, Jiaran Zhou, and Junyu Dong.
\newblock Forensics adapter: Adapting clip for generalizable face forgery
  detection.
\newblock In \emph{2025 IEEE/CVF Conference on Computer Vision and Pattern
  Recognition (CVPR)}, pages 19207--19217, 2025.
\newblock \doi{10.1109/CVPR52734.2025.01789}.

\bibitem[De~Min et~al.(2023)De~Min, Mancini, Alahari, Alameda-Pineda, and
  Ricci]{demin2023effectiveness}
Thomas De~Min, Massimiliano Mancini, Karteek Alahari, Xavier Alameda-Pineda,
  and Elisa Ricci.
\newblock On the effectiveness of layernorm tuning for continual learning in
  vision transformers.
\newblock In \emph{2023 IEEE/CVF International Conference on Computer Vision
  Workshops (ICCVW)}, pages 3577--3586, 2023.
\newblock \doi{10.1109/ICCVW60793.2023.00385}.

\bibitem[Dolhansky et~al.(2020)Dolhansky, Bitton, Pflaum, Lu, Howes, Wang, and
  Ferrer]{dolhansky2020deepfake}
Brian Dolhansky, Joanna Bitton, Ben Pflaum, Jikuo Lu, Russ Howes, Menglin Wang,
  and Cristian~Canton Ferrer.
\newblock The deepfake detection challenge (dfdc) dataset, June 2020.

\bibitem[Du et~al.(2026)Du, Ma, Zhu, Yang, Niu, Qu, Fang, Wang, Liu, Liu, and
  Zhou]{du2026can}
Bo~Du, Xiaochen Ma, Xuekang Zhu, Zhe Yang, Chaoqun Niu, Chenfan Qu, Mingqi
  Fang, Zhenming Wang, Jingjing Liu, Jian Liu, and Ji-Zhe Zhou.
\newblock Can we build a monolithic model for fake image detection? sica:
  Semantic-induced constrained adaptation for unified-yet-discriminative
  artifact feature space reconstruction.
\newblock February 2026.
\newblock \doi{10.48550/ARXIV.2602.06676}.

\bibitem[Dufour and Gully(2019)]{dufour2019contributing}
Nick Dufour and Andrew Gully.
\newblock Contributing data to deepfake detection research.
\newblock Google Research Blog, September 2019.
\newblock URL
  \url{https://research.google/blog/contributing-data-to-deepfake-detection-research/}.
\newblock Accessed: \today.

\bibitem[Giannou et~al.(2023)Giannou, Rajput, and
  Papailiopoulos]{giannou2023expressive}
Angeliki Giannou, Shashank Rajput, and Dimitris Papailiopoulos.
\newblock The expressive power of tuning only the normalization layers.
\newblock In Gergely Neu and Lorenzo Rosasco, editors, \emph{Proceedings of
  Thirty Sixth Conference on Learning Theory}, volume 195 of \emph{Proceedings
  of Machine Learning Research}, pages 4130--4131. PMLR, 12--15 Jul 2023.
\newblock URL \url{https://proceedings.mlr.press/v195/giannou23a.html}.

\bibitem[Hou et~al.(2024)Hou, Fu, Chen, Li, Zhang, and
  Zhao]{hou2024polyglotfake}
Yang Hou, Haitao Fu, Chunkai Chen, Zida Li, Haoyu Zhang, and Jianjun Zhao.
\newblock Polyglotfake: A novel multilingual and multimodal deepfake dataset.
\newblock In \emph{Pattern Recognition: 27th International Conference, ICPR
  2024, Kolkata, India, December 1–5, 2024, Proceedings, Part XIV}, page
  180–193, Berlin, Heidelberg, 2024. Springer-Verlag.
\newblock ISBN 978-3-031-78340-1.
\newblock \doi{10.1007/978-3-031-78341-8_12}.

\bibitem[Hu et~al.(2022)Hu, Shen, Wallis, Allen-Zhu, Li, Wang, Wang, and
  Chen]{hu2022lora}
Edward~J Hu, Yelong Shen, Phillip Wallis, Zeyuan Allen-Zhu, Yuanzhi Li, Shean
  Wang, Lu~Wang, and Weizhu Chen.
\newblock Lo{RA}: Low-rank adaptation of large language models.
\newblock In \emph{International Conference on Learning Representations}, 2022.
\newblock URL \url{https://openreview.net/forum?id=nZeVKeeFYf9}.

\bibitem[Jia et~al.(2022)Jia, Li, and Lyu]{jia2022model}
Shan Jia, Xin Li, and Siwei Lyu.
\newblock Model attribution of face-swap deepfake videos.
\newblock In \emph{2022 IEEE International Conference on Image Processing
  (ICIP)}, pages 2356--2360, 2022.
\newblock \doi{10.1109/ICIP46576.2022.9897972}.

\bibitem[Jiang et~al.(2020)Jiang, Li, Wu, Qian, and
  Loy]{jiang2020deeperforensics}
Liming Jiang, Ren Li, Wayne Wu, Chen Qian, and Chen~Change Loy.
\newblock Deeperforensics-1.0: A large-scale dataset for real-world face
  forgery detection.
\newblock In \emph{2020 IEEE/CVF Conference on Computer Vision and Pattern
  Recognition (CVPR)}, pages 2886--2895. IEEE, June 2020.
\newblock \doi{10.1109/cvpr42600.2020.00296}.

\bibitem[Kendall et~al.(2018)Kendall, Gal, and Cipolla]{kendall2018multi}
Alex Kendall, Yarin Gal, and Roberto Cipolla.
\newblock Multi-task learning using uncertainty to weigh losses for scene
  geometry and semantics.
\newblock In \emph{Proceedings of the IEEE Conference on Computer Vision and
  Pattern Recognition (CVPR)}, June 2018.
\newblock URL
  \url{https://openaccess.thecvf.com/content_cvpr_2018/html/Kendall_Multi-Task_Learning_Using_CVPR_2018_paper.html}.

\bibitem[Khalid et~al.(2021)Khalid, Tariq, Kim, and Woo]{khalid2021fakeavceleb}
Hasam Khalid, Shahroz Tariq, Minha Kim, and Simon~S. Woo.
\newblock Fake{AVC}eleb: A novel audio-video multimodal deepfake dataset.
\newblock In \emph{Thirty-fifth Conference on Neural Information Processing
  Systems Datasets and Benchmarks Track (Round 2)}, 2021.
\newblock URL \url{https://openreview.net/forum?id=TAXFsg6ZaOl}.

\bibitem[Kwon et~al.(2021)Kwon, You, Nam, Park, and Chae]{kwon2021kodf}
Patrick Kwon, Jaeseong You, Gyuhyeon Nam, Sungwoo Park, and Gyeongsu Chae.
\newblock Kodf: A large-scale korean deepfake detection dataset.
\newblock In \emph{2021 IEEE/CVF International Conference on Computer Vision
  (ICCV)}, pages 10724--10733, 2021.
\newblock \doi{10.1109/ICCV48922.2021.01057}.

\bibitem[Li et~al.(2020)Li, Yang, Sun, Qi, and Lyu]{li2020celeb}
Yuezun Li, Xin Yang, Pu~Sun, Honggang Qi, and Siwei Lyu.
\newblock Celeb-{DF}: {A} {Large}-{Scale} {Challenging} {Dataset} for
  {DeepFake} {Forensics}.
\newblock In \emph{2020 {IEEE}/{CVF} {Conference} on {Computer} {Vision} and
  {Pattern} {Recognition} ({CVPR})}, pages 3204--3213, June 2020.
\newblock \doi{10.1109/CVPR42600.2020.00327}.
\newblock URL \url{https://ieeexplore.ieee.org/abstract/document/9156368}.

\bibitem[Li et~al.(2025)Li, Zhu, Cui, and Lyu]{li2025celeb}
Yuezun Li, Delong Zhu, Xinjie Cui, and Siwei Lyu.
\newblock Celeb-{DF}++: {A} {Large}-scale {Challenging} {Video} {DeepFake}
  {Benchmark} for {Generalizable} {Forensics}, July 2025.

\bibitem[Nguyen et~al.(2025)Nguyen, Astrid, Kacem, Ghorbel, and
  Aouada]{nguyen2025vulnerability}
Dat Nguyen, Marcella Astrid, Anis Kacem, Enjie Ghorbel, and Djamila Aouada.
\newblock Vulnerability-aware spatio-temporal learning for generalizable
  deepfake video detection.
\newblock In \emph{2025 IEEE/CVF International Conference on Computer Vision
  (ICCV)}, pages 10786--10796. IEEE, October 2025.
\newblock \doi{10.1109/iccv51701.2025.01004}.

\bibitem[Rössler et~al.(2019)Rössler, Cozzolino, Verdoliva, Riess, Thies, and
  Nießner]{roessler2019faceforensics++}
Andreas Rössler, D.~Cozzolino, L.~Verdoliva, C.~Riess, Justus Thies, and
  M.~Nießner.
\newblock {FaceForensics++}: Learning to detect manipulated facial images.
\newblock In \emph{IEEE International Conference on Computer Vision}, pages
  1--11. IEEE, October 2019.
\newblock \doi{10.1109/ICCV.2019.00009}.

\bibitem[Siméoni et~al.(2025)Siméoni, Vo, Seitzer, Baldassarre, Oquab, Jose,
  Khalidov, Szafraniec, Yi, Ramamonjisoa, Massa, Haziza, Wehrstedt, Wang,
  Darcet, Moutakanni, Sentana, Roberts, Vedaldi, Tolan, Brandt, Couprie,
  Mairal, Jégou, Labatut, and Bojanowski]{simeoni2025dinov3}
Oriane Siméoni, Huy~V. Vo, Maximilian Seitzer, Federico Baldassarre, Maxime
  Oquab, Cijo Jose, Vasil Khalidov, Marc Szafraniec, Seungeun Yi, Michaël
  Ramamonjisoa, Francisco Massa, Daniel Haziza, Luca Wehrstedt, Jianyuan Wang,
  Timothée Darcet, Théo Moutakanni, Leonel Sentana, Claire Roberts, Andrea
  Vedaldi, Jamie Tolan, John Brandt, Camille Couprie, Julien Mairal, Hervé
  Jégou, Patrick Labatut, and Piotr Bojanowski.
\newblock {DINOv3}, August 2025.

\bibitem[Wang and Isola(2020)]{wang2020understanding}
Tongzhou Wang and Phillip Isola.
\newblock Understanding contrastive representation learning through alignment
  and uniformity on the hypersphere.
\newblock In Hal~Daumé III and Aarti Singh, editors, \emph{Proceedings of the
  37th International Conference on Machine Learning}, volume 119 of
  \emph{Proceedings of Machine Learning Research}, pages 9929--9939. PMLR,
  13--18 Jul 2020.
\newblock URL \url{https://proceedings.mlr.press/v119/wang20k.html}.

\bibitem[Xu et~al.(2024)Xu, Chen, Song, Han, Shan, and Jiang]{xu2024identity}
Junhao Xu, Jingjing Chen, Xue Song, Feng Han, Haijun Shan, and Yu-Gang Jiang.
\newblock Identity-driven multimedia forgery detection via reference
  assistance.
\newblock In \emph{Proceedings of the 32nd ACM International Conference on
  Multimedia}, MM '24, page 3887–3896, New York, NY, USA, 2024. Association
  for Computing Machinery.
\newblock ISBN 9798400706868.
\newblock \doi{10.1145/3664647.3680622}.

\bibitem[Yan et~al.(2023)Yan, Zhang, Yuan, Lyu, and Wu]{yan2023deepfakebench}
Zhiyuan Yan, Yong Zhang, Xinhang Yuan, Siwei Lyu, and Baoyuan Wu.
\newblock Deepfakebench: a comprehensive benchmark of deepfake detection.
\newblock In \emph{Proceedings of the 37th International Conference on Neural
  Information Processing Systems}, NIPS '23, Red Hook, NY, USA, 2023. Curran
  Associates Inc.
\newblock \doi{10.5555/3666122.3666323}.

\bibitem[Yan et~al.(2024)Yan, Luo, Lyu, Liu, and Wu]{yan2024transcending}
Zhiyuan Yan, Yuhao Luo, Siwei Lyu, Qingshan Liu, and Baoyuan Wu.
\newblock Transcending forgery specificity with latent space augmentation for
  generalizable deepfake detection.
\newblock In \emph{2024 IEEE/CVF Conference on Computer Vision and Pattern
  Recognition (CVPR)}, pages 8984--8994. IEEE, June 2024.
\newblock \doi{10.1109/cvpr52733.2024.00858}.

\bibitem[Yan et~al.(2025)Yan, Wang, Jin, Zhang, Liu, Chen, Yao, Ding, Wu, and
  Yuan]{yan2025orthogonal}
Zhiyuan Yan, Jiangming Wang, Peng Jin, Ke-Yue Zhang, Chengchun Liu, Shen Chen,
  Taiping Yao, Shouhong Ding, Baoyuan Wu, and Li~Yuan.
\newblock Orthogonal subspace decomposition for generalizable ai-generated
  image detection.
\newblock In \emph{Proceedings of the 42nd International Conference on Machine
  Learning}, ICML'25, Vancouver, Canada, 2025. JMLR.org.
\newblock \doi{10.5555/3780338.3783150}.

\bibitem[Yang et~al.(2019)Yang, Li, and Lyu]{yang2019exposing}
Xin Yang, Yuezun Li, and Siwei Lyu.
\newblock Exposing deep fakes using inconsistent head poses.
\newblock In \emph{ICASSP 2019 - 2019 IEEE International Conference on
  Acoustics, Speech and Signal Processing (ICASSP)}, pages 8261--8265, 2019.
\newblock \doi{10.1109/ICASSP.2019.8683164}.

\bibitem[Yermakov et~al.(2026)Yermakov, Cech, Matas, and
  Fritz]{yermakov2026deepfake}
Andrii Yermakov, Jan Cech, Jiri Matas, and Mario Fritz.
\newblock Deepfake detection that generalizes across benchmarks.
\newblock In \emph{2026 IEEE/CVF Winter Conference on Applications of Computer
  Vision (WACV)}, pages 773--783. IEEE, March 2026.
\newblock \doi{10.1109/WACV61042.2026.00082}.

\bibitem[Zhou et~al.(2021)Zhou, Wang, Liang, and Shen]{zhou2021face}
Tianfei Zhou, Wenguan Wang, Zhiyuan Liang, and Jianbing Shen.
\newblock Face forensics in the wild.
\newblock In \emph{2021 IEEE/CVF Conference on Computer Vision and Pattern
  Recognition (CVPR)}, pages 5774--5784. IEEE, June 2021.
\newblock \doi{10.1109/cvpr46437.2021.00572}.

\end{thebibliography}
\end{document}